\documentclass{article}

\usepackage[preprint]{neurips_2025}
\usepackage{booktabs}
\usepackage{amsmath}
\usepackage{amssymb} 
\usepackage{graphicx}
\usepackage{caption}
\usepackage[table]{xcolor}
\usepackage{amsmath} 
\usepackage{amssymb} 
\usepackage{booktabs}
\colorlet{colorHuman}{cyan!10}
\colorlet{colorSelected}{green!10}
\colorlet{colorDirect}{yellow!15}
\colorlet{colorOurs}{magenta!10}
\colorlet{colorPattern}{orange!10}

\usepackage[utf8]{inputenc} 
\usepackage[T1]{fontenc}    
\usepackage{hyperref}       
\usepackage{url}            
\usepackage{booktabs}       
\usepackage{amsmath}
\usepackage{amsfonts}       
\usepackage{nicefrac}       
\usepackage{microtype}      
\usepackage{xcolor}         
\usepackage{amsfonts}
\usepackage{algorithm}
\usepackage{algorithmic}
\usepackage{graphicx}
\usepackage{multirow}
\usepackage{multicol}
\usepackage{changepage}
\usepackage{enumitem}
\usepackage{tcolorbox}
\usepackage{subfigure}
\usepackage{graphicx}
\usepackage{subcaption}

\newcounter{promptcounter}

\title{Synthetic Data RL: Task Definition Is All You Need}

\author{%
Yiduo Guo \\
Peking University \\
\And
\hspace{-.09cm}Zhen Guo \\
\hspace{-.09cm}MIT \\
\And
\hspace{.14cm}Chuanwei Huang \\
\hspace{.14cm}Peking University \\
\And
Zi-Ang Wang \\
Peking University \\
\AND
\hspace{-.14cm}Zekai Zhang \\
\hspace{-.14cm}Peking University \\
\And
\hspace{-.4cm}Haofei Yu \\
\hspace{-.4cm}UIUC \\
\And
Huishuai Zhang \\
Peking University \\
\And
Yikang Shen \\
MIT-IBM \\
}

\begin{document}

\maketitle

\begin{abstract}
Reinforcement learning (RL) is a powerful way to adapt foundation models to specialized tasks, but its reliance on large-scale human-labeled data limits broad adoption. 
We introduce \textbf{Synthetic Data RL}, a simple and general framework that reinforcement fine-tunes models using \textbf{only} synthetic data generated from a task definition. 
Our method first generates question and answer pairs from the task definition and retrieved documents, then adapts the difficulty of the question based on model solvability, and selects questions using the average pass rate of the model across samples for RL training. 
On Qwen-2.5-7B, our method achieves a 29.2\% absolute improvement over the base model on GSM8K (+2.9 pp vs. instruction-tuned, +6.6 pp vs. Self-Instruct), 8.7\% on MATH, 13.1\% on GPQA (+7.0 pp vs. SynthLLM), 8.9\% on MedQA, 17.7 \% on CQA (law) and 13.7\% on CFA (finance).
It surpasses supervised fine-tuning under the same data budget and nearly matches RL with full human data across datasets (e.g., +17.2 pp on GSM8K). Adding 100 human demonstrations improves the performance of GSM8K only by 0.4 pp, showing a limited added value. By reducing human data annotation, Synthetic Data RL enables scalable and efficient RL-based model adaptation. Code and demos are available at \url{https://github.com/gydpku/Data_Synthesis_RL/}. 

\end{abstract}

\begin{figure}[!ht]
    \centering
    \includegraphics[width=0.98\linewidth]{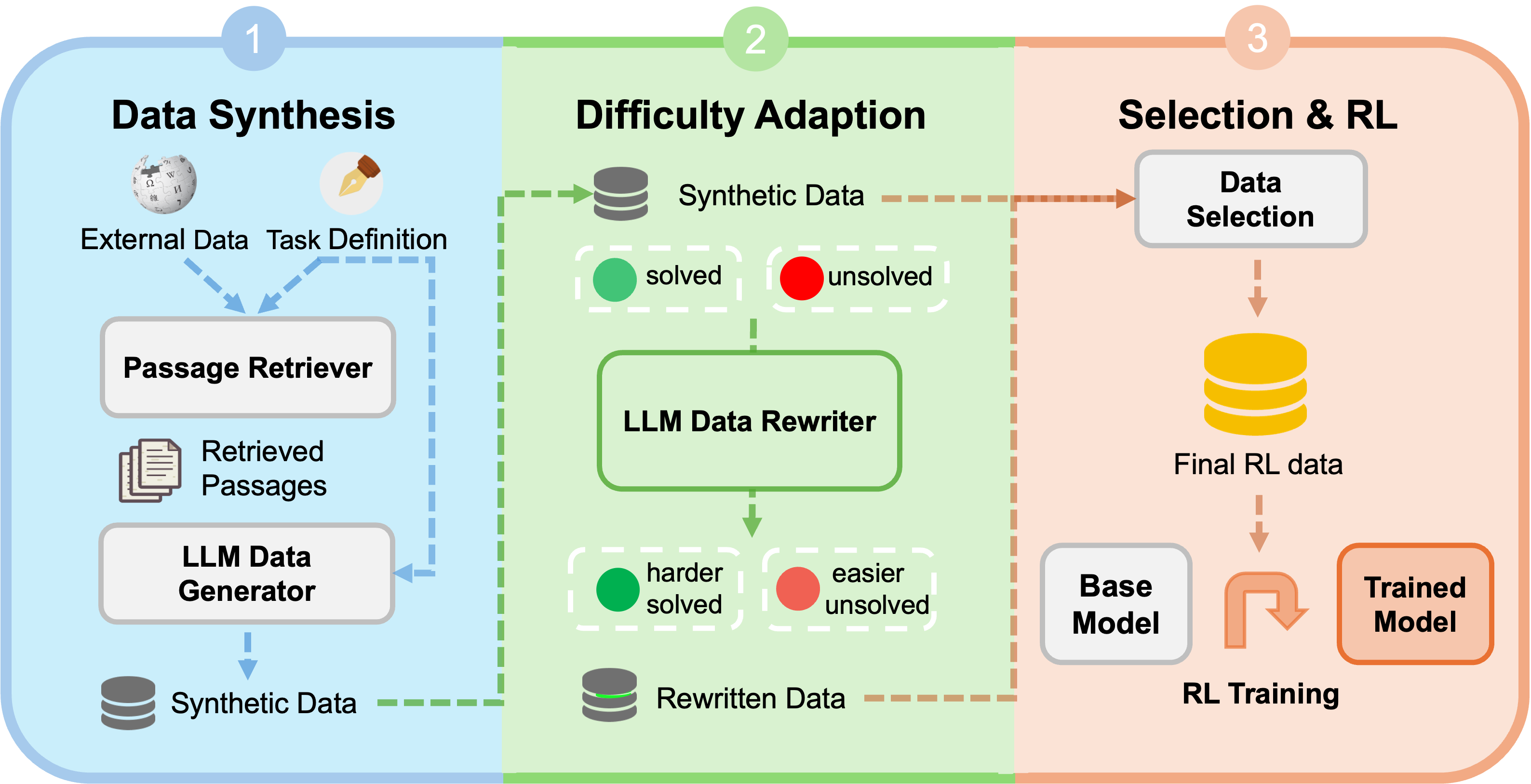}
    \vspace{2mm}
    \caption{High-level overview for Synthetic Data RL.}
    \label{fig.overview}
\end{figure}

\section{Introduction}

Foundation models~\cite{openai2024gpt4technicalreport, geminiteam2024geminifamilyhighlycapable, grattafiori2024llama3herdmodels, gunter2024appleintelligencefoundationlanguage, zhong2024evaluationopenaio1opportunities,qwen2.5, deepseekai2025deepseekr1incentivizingreasoningcapability} have set new standards in general-purpose language understanding. However, these models often underperform in specialized domains such as math, medicine, law, and finance, where domain-specific knowledge are essential~\cite{singhal2023expertlevelmedicalquestionanswering, Lee_2025, ke2025demystifyingdomainadaptiveposttrainingfinancial}. Adapting these models requires large-scale human-labeled data~\cite{dubois2024alpacafarmsimulationframeworkmethods}, which is costly, slow, and often impractical in real applications.

Efforts in lightweight adaptation, like retrieval-agmented generation (RAG)~\cite{zhang2024raftadaptinglanguagemodel, rakin2024leveragingdomainadaptationretrieval, cheng2025surveyknowledgeorientedretrievalaugmentedgeneration, lee-etal-2025-trag}, offer a non-parametric alternative by incorporating external knowledge at inference time. Although effective in some scenarios, RAG struggles to leverage the latent reasoning capabilities of the model~\cite{ke2025naacl2025tutorialadaptationlarge, yue2025doesreinforcementlearningreally}.

To address these challenges, we propose \textbf{Synthetic Data RL}, a simple and general framework for adapting foundation models using \emph{only synthetic data} generated from minimal human input. Starting from a task definition, our method synthesizes diverse, domain-specific examples and fine-tunes the model using reinforcement learning (RL). This enables parametric adaptation that embeds domain knowledge directly into the model, without requiring any human-labeled data. While methods like Reinforcement Fine-Tuning (RFT)~\cite{openai2024rft} and OpenRFT~\cite{zhang2024openrftadaptingreasoningfoundation} have demonstrated the potential of RL for model adaptation, our framework distinguishes itself by relying on synthetic data generated from a single task definition.

The \textbf{Synthetic Data RL} comprises three key steps (also see in Figure~\ref{fig.overview}):

\begin{enumerate}
\item \textbf{Knowledge-Guided Synthesis:} We combine retrieved external knowledge with task-specific patterns extracted from minimal input to generate synthetic examples that are both grounded and aligned with the target task~\cite{viswanathan2023prompt2modelgeneratingdeployablemodels, divekar-durrett-2024-synthesizrr, zhu2025barecombiningbaseinstructiontuned, goldie2025syntheticdatageneration}.
\item \textbf{Difficulty-Adaptive Curriculum:} Model feedback is used to adjust the complexity of generated samples, balancing the difficulty of the dataset to avoid too easy or too hard examples~\cite{DART-Math}.
\item \textbf{High-Potential Sample Selection \& RL:} We select examples where the model shows partial understanding and fine-tune on them using reinforcement learning (e.g., GRPO), reinforcing generalizable behavior~\cite{dubois2024alpacafarmsimulationframeworkmethods, ramesh2024grouprobustpreferenceoptimization}.

\end{enumerate}

We investigate the effectiveness of our method and summarize our results below:
\begin{itemize}
\item It achieves \textbf{91.7\%} on \textbf{GSM8K}, surpassing all baselines including the official instruction-tuned model (88.8\%), and generalizes well across domains like \textbf{MATH} (72.0\%), \textbf{LogiQA} (55.0\%), \textbf{MedQA} (medical, 64.5\%), \textbf{CQA} (law, 92.4\%) and \textbf{CFA} (finance, 73.2\%) (see Table~\ref{table.main}).
\item It matches or exceeds RL with full human-annotated data and consistently outperforms supervised fine-tuning under the same data budget (Sec.~\ref{sec.main}).
\item It gains only marginal benefit from 100 extra human demonstrations (e.g., GSM8K: 91.7\% $\rightarrow$ 92.1\%) (Sec.~\ref{sec.main}).
\item It enables an instructor model (e.g., Qwen-2.5-7B-Instruct) to fine-tune its own base model and produce an even stronger model (Sec.~\ref{sec.understanding}).

\item It may require a model with behaviors such as verification and backtracking (Sec.~\ref{sec.understanding}).
\end{itemize}

\section{Related works}
\subsection{Synthetic Data Generation with LLMs}

Synthetic data from large language models (LLMs) now drives advances in reasoning, tool use, multimodal learning, and alignment. Early works such as Alpaca and Vicuna~\cite{alpaca,vicuna2023} showed that synthetic data can elicit strong instruction-following in 7B models. For reasoning tasks, pipelines like WizardMath~\cite{luo2025wizardmathempoweringmathematicalreasoning}, MetaMath~\cite{yu2024metamathbootstrapmathematicalquestions}, and AlphaGeometry~\cite{53097} generate and verify math problems, while code agents such as CodeRL~\cite{le2022coderlmasteringcodegeneration} and WizardCoder~\cite{luo2023wizardcoderempoweringcodelarge} use execution-based filtering to increase accuracy~\cite{luo2023wizardcoderempoweringcodelarge,yu2024metamathbootstrapmathematicalquestions,trinh2024solving}. In tool use and planning, models like Toolformer, ToolAlpaca, and Voyager learn long-horizon control via synthetic trajectories~\cite{schick2023toolformerlanguagemodelsteach,tang2023toolalpacageneralizedtoollearning,wang2023voyageropenendedembodiedagent}. For multimodal grounding, render-and-describe pipelines such as Pix2Struct, LLaVA, and MatCha create fine-grained image–text pairs that outperform noisy web captions~\cite{lee2023pix2structscreenshotparsingpretraining,liu2023visualinstructiontuning,liu2023matchaenhancingvisuallanguage}. In multilingual and alignment settings, back-translation and synthetic QA (e.g., PaxQA) improve generalization~\cite{sennrich2016improvingneuralmachinetranslation,li2023paxqageneratingcrosslingualquestion}, while Self-Instruct and Constitutional AI methods reduce dependence on human-labeled feedback~\cite{wang2023selfinstructaligninglanguagemodels, bai2022constitutionalaiharmlessnessai}. Best practices emphasize the need for factuality and bias filtering~\cite{feng2023factkbgeneralizablefactualityevaluation}, prioritizing quality over quantity~\cite{chen2024alpagasustrainingbetteralpaca}. Open challenges remain in defining scaling laws~\cite{qin2025scalinglawssyntheticdata} and enabling self-improving data loops~\cite{yuan2025selfrewardinglanguagemodels}.

\subsection{Reinforcement Learning with Synthetic Data}

Reinforcement learning aligns LLMs using feedback signals. Standard methods such as RLHF with PPO~\cite{ouyang2022traininglanguagemodelsfollow,schulman2017proximalpolicyoptimizationalgorithms} and RLAIF~\cite{bai2022constitutionalaiharmlessnessai} use a reward model and a critic, but are complex and resource intensive. Simpler approaches like DPO~\cite{rafailov2024directpreferenceoptimizationlanguage} and GRPO~\cite{deepseek-math} skip the critic and learn directly from preference pairs. REINFORCE and RLOO keeps the reward model but removes the critic, using basic policy gradients to match or exceed PPO performance with lower compute~\cite{ahmadian2024basicsrevisitingreinforcestyle}. Recent approaches shift from response-level to step-level feedback by using synthetic trajectories. SWiRL~\cite{goldie2025syntheticdatageneration}, DQO~\cite{ji2025enhancingmultistepreasoningabilities}, and OREO~\cite{wang2024offlinereinforcementlearningllm} label individual reasoning/tool-use actions, supporting long-horizon planning without gold answers. Verifiable signals, such as coding pass rates (RLEF~\cite{gehring2025rlefgroundingcodellms}) or math checkers (Tulu-3~\cite{lambert2025tulu3pushingfrontiers}), further reduce reliance on human labels. 


Scaling synthetic RL relies on reward fidelity and data diversity~\cite{Lu2023SyntheticER}. Large-scale pipelines such as SYNTHETIC-1~\cite{2025synthetic1}, STaR~\cite{zelikman2022starbootstrappingreasoningreasoning}, and ReSTEM~\cite{singh2024humandatascalingselftraining} generate millions of filtered samples. Efficiency improvements—negative-sample training~\cite{setlur2024rlincorrectsyntheticdata}, teacher–student distillation~\cite{distilabel-argilla-2024}, and majority-vote self-verification~\cite{zuo2025ttrltesttimereinforcementlearning}—reduce compute cost. Resource-efficient recipes such as SimpleRL~\cite{zeng2025simplerlzooinvestigatingtamingzero} and TinyZero~\cite{tinyzero} show that even small base models can generalize better than supervised fine-tuning~\cite{chu2025sftmemorizesrlgeneralizes, sun2025climbingladderreasoningllms}, using a few hundred epochs with reliable synthetic rewards. Though there is ongoing debate about whether it truly expands reasoning capabilities beyond the base model~\cite{yue2025doesreinforcementlearningreally}. Concurrent studies also find that RL can achieve strong performance with minimal data, such as a single training example~\cite{wang2025reinforcementlearningreasoninglarge}, or even without external data through self-play mechanisms~\cite{zhao2025absolutezeroreinforcedselfplay}, but limit on math and code evaluations.

\section{Problem Definition}

Given a task, the machine learning objective is formulated as the following optimization problem:
\begin{equation}
\min_{\theta} \; \mathcal{L}(\mathcal{A}, \mathcal{D}, \mathcal{M}_{\text{base}}, \mathcal{H}),
\end{equation}
where $\mathcal{L}$ denotes the task loss, $\mathcal{A}$ is the learning algorithm (e.g., reinforcement learning (RL) or supervised fine-tuning (SFT)), $\mathcal{D}$ is the task-specific data, $\mathcal{M}_{\text{base}}$ is the initial model with rich prior knowledge, and $\mathcal{H}$ represents human involvement, which may include providing task descriptions $\mathcal{I}$, annotating data, or giving feedback during training.

While human input can be crucial for effective learning, our goal is to minimize the reliance on human annotation and intervention, while maintaining—or even improving—the final task performance. To achieve this, we propose a method that combines our automatic data synthesis method with RL algorithm. This approach reduces human effort to only providing the task description, significantly lowering the cost of human supervision.
\section{Our algorithm}
In this section, we introduce our synthetic data reinforcement learning (RL) framework designed to effectively train a base language model $\mathcal{M}_{\text{base}}$. Our method leverages an instructor model to systematically generate, and adapt synthetic data.  It then uses the base model to select synthetic training samples that maximize learning potential and automatically conducts RL training to train the resulting model $\mathcal{M}_{\textit{trained}}$. Specifically, the framework comprises four key components: 
\begin{enumerate}
    \item \textbf{Passage Retriever $\mathcal{P}$}: taking keywords as input and uses a retrieval algorithm to find relevant passages from a large collection of high-quality text passages  $\mathrm{L}$, such as Wikipedia.

    \item \textbf{LLM Data Generator} ($\text{LLM}_{\text{generator}}$): taking task instructions, demo examples (optional), and passages as input and uses a powerful instructor language model ($\mathrm{I}$) to create new training data. To ensure the quality of the generated outputs, a verification step is performed—typically by sampling multiple responses for each input and applying a consensus mechanism such as \emph{majority voting} to select the most frequent output as the final estimated output.

    \item \textbf{LLM Data Re-writer} ($\text{LLM}_{\text{writer}}$): taking synthetic data as input and uses the instructor language model ($\mathrm{I}$) to modify the difficulty of them, outputting a harder version or simpler version. Similar to the data generation step, it also verifies the quality of these rewritten examples.

    \item \textbf{Trainer ($T$)}: module that uses RL to train $\mathcal{M}_{\text{base}}$ with synthesis training examples.

\end{enumerate}

In our method, the user is required to provide a task definition consisting of three components: a task description instruction ($\mathcal{I}_{\textit{des}}$), an input format instruction ($\mathcal{I}_{\textit{input}}$), and an output format instruction ($\mathcal{I}_{\textit{output}}$), as illustrated in Figure~\ref{fig:instruction}. Given this definition, our method follows the procedure below to train the resulting model $\mathcal{M}_{\textit{trained}}$.

\subsection{External Knowledge-Guided Synthesis}

\paragraph{Keyword extraction and relevant passage retrieval:} Initially, we consider a relevant passage retrieval step, which acts as a knowledge augmentation strategy to provide the LLM generator with broader contextual information in the data synthesis process. Specifically, given a task description instruction $\mathcal{I}_{\textit{des}}$ and some demonstration examples $\mathcal{D}_\text{example}$ (optional), we use an instructor model $\mathrm{I}$ to derive a set of domain-specific keywords $\mathcal{K}$ (see the prompt in Figure~\ref{fig:keyword}). These keywords serve as an intermediate representation for identifying pertinent information within an external passage library $\mathrm{L}$, from which we retrieve a collection of relevant passages $\mathcal{R}$:
\begin{equation} \label{eq:keyword_retrieval}
\begin{aligned}
\mathcal{K} &= \mathrm{I}(\mathcal{D}_\text{example}, \mathcal{I}_{\textit{des}}) \\
\mathcal{R} &= \mathcal{P}(\mathcal{K}, \mathrm{L})
\end{aligned}
\end{equation}

\begin{table}[ht]
\centering
\caption{Diversity comparison across examples. Human-created examples are sampled from the Algebra task training data in the MATH dataset. We selected one demo example from this dataset to guide generation. One approach, yielding 'Examples generated directly', used this demo example exclusively for guidance. Our proposed approach, resulting in 'Examples using our method', involved first inducing a pattern and then using both the pattern and the demo example for data generation.}
\label{tab:example_comparison}
\resizebox{0.98\textwidth}{!}{%
\begin{tabular}{ll}
\toprule
\rowcolor{colorHuman} & \textbf{Example 1:} \textit{Input:} What is the degree of the polynomial $\bigl(4 + 5x^{3} + 100 + 2\pi x^{4} + \sqrt{10}\,x^{4} + 9\bigr)$? \\
\rowcolor{colorHuman} & \quad\quad\quad\quad\quad \textit{Output:} ... (CoT thinking process) Answer:\quad $\boxed{4}$ \\
\rowcolor{colorHuman} & \\ 
\rowcolor{colorHuman} & \textbf{Example 2:} \textit{Input:} If $x = 2$ and $y = 5$, then what is the value of $\frac{x^4+2y^2}{6}$ ? \\
\rowcolor{colorHuman} {\large\textbf{Human-created examples}} & \quad\quad\quad\quad\quad \textit{Output:} ... (CoT thinking process) Answer:\quad $\boxed{11}$ \\
\rowcolor{colorHuman} & \\ 
\rowcolor{colorHuman} & \textbf{Example 3:} \textit{Input:} Let $f(x) =
 \begin{cases}
 x/2 &\quad \text{if } x \text{ is even}, \\
 3x+1 &\quad \text{if } x \text{ is odd}.
 \end{cases}
 $
 What is $f(f(f(f(1))))$? \\
\rowcolor{colorHuman} & \quad\quad\quad\quad\quad \textit{Output:} ... (CoT thinking process) Answer:\quad $\boxed{4}$ \\
\midrule
\rowcolor{colorSelected} {\large\textbf{Demo example}} &  \textbf{Example : }  \textit{Input:} What is the degree of the polynomial $\bigl(4 + 5x^{3} + 100 + 2\pi x^{4} + \sqrt{10}\,x^{4} + 9\bigr)$? \\
\rowcolor{colorSelected} & \quad\quad\quad\quad\quad \textit{Output:} ... (CoT thinking process) Answer:\quad $\boxed{4}$ \\
\midrule
\rowcolor{colorDirect} & \textbf{Example 1:} \textit{Input:} What is the degree of the polynomial $\bigl(3x^{2} + 7 - x^{5} + 2x^{2} + 10x^{5} + \sqrt{2}\bigr)$? \\
\rowcolor{colorDirect} & \quad\quad\quad\quad\quad \textit{Output:} ... (CoT thinking process) Answer:\quad $\boxed{5}$ \\
\rowcolor{colorDirect} & \\ 
\rowcolor{colorDirect} & \textbf{Example 2:} \textit{Input:} What is the degree of the polynomial $\bigl(10y + 8y^{6} - 3y + y^{6} + 5y^{2} + 1\bigr)$? \\
\rowcolor{colorDirect} {\large\textbf{Examples generated directly}} & \quad\quad\quad\quad\quad \textit{Output:} ... (CoT thinking process) Answer:\quad $\boxed{6}$ \\
\rowcolor{colorDirect} & \\ 
\rowcolor{colorDirect} & \textbf{Example 3:} \textit{Input:} What is the degree of the polynomial $\bigl(z^{3} - z^{3} + z^{2} + 5z^{2} + \pi z + 8\bigr)$? \\
\rowcolor{colorDirect} & \quad\quad\quad\quad\quad \textit{Output:} ... (CoT thinking process) Answer:\quad $\boxed{2}$ \\
\midrule
\rowcolor{colorPattern}  & The task involves solving algebraic problems by applying mathematical knowledge and reasoning. \\  \rowcolor{colorPattern} & The input consists of a math problem, often involving equations, functions, or geometric properties. \\  \rowcolor{colorPattern} &  The output requires a step-by-step explanation leading to the final answer, which is presented within \\ \rowcolor{colorPattern} {\large\textbf{Pattern}} &  a boxed format. The problems may involve operations such as factoring, solving equations, \\ \rowcolor{colorPattern} & finding intercepts, evaluating expressions, or simplifying complex numbers. The solutions often \\ \rowcolor{colorPattern} &  involve algebraic manipulation, use of formulas, and logical reasoning to arrive at the correct answer.\\  
\midrule
\rowcolor{colorOurs} & \textbf{Example 1:} \textit{Input:} Solve for $x$ in the equation $3(x+2) - 2(2x - 1) = 4 + 5x$. \\
\rowcolor{colorOurs} & \quad\quad\quad\quad\quad \textit{Output:} ... (CoT thinking process) Answer:\quad $\boxed{\frac{2}{3}}$ \\ 
\rowcolor{colorOurs} & \\ 
\rowcolor{colorOurs} & \textbf{Example 2:} \textit{Input:} In a principal $G$-bundle over a space $X$, the characteristic class can be\\
\rowcolor{colorOurs} & represented by a cohomology class. If a characteristic class $c(P)$ is associated with a\\
\rowcolor{colorOurs} & cohomology class of degree 2, and the total dimension of the manifold M is 4, what is \\
\rowcolor{colorOurs} & the degree of the characteristic number obtained by pairing this class with the manifold? \\
\rowcolor{colorOurs} {\large\textbf{Examples using our method}} & \quad\quad\quad\quad\quad \textit{Output:} ... (CoT thinking process) Answer:\quad $\boxed{4}$ \\
\rowcolor{colorOurs} & \\ 
\rowcolor{colorOurs} & \textbf{Example 3:} \textit{Input:} Given that the modular form $f$ associated with an elliptic curve $E$ over $\mathbb{Q}$ \\
\rowcolor{colorOurs} & satisfies $f(\tau) = \sum_{n=1}^{\infty} a_n q^n$, where $q = e^{2\pi i \tau}$, and the Fourier coefficients $a_p = 1$ for all \\
\rowcolor{colorOurs} & primes $p$, determine the value of $a_{25}$. \\
\rowcolor{colorOurs} & \quad\quad\quad\quad\quad \textit{Output:} ... (CoT thinking process) Answer:\quad $\boxed{-4}$\\ 
\bottomrule
\end{tabular}
}
\end{table}

\paragraph{Data generation with sample pattern summarization:} 
Subsequently, we leverage the LLM generator to synthesize an initial set of $N$ task samples $\mathcal{S}_{\text{initial}} = \{s_1, s_2, ..., s_{N}\}$, conditioned on the retrieved passages $\mathcal{R}$, and three task instructions (See prompt in~\ref{fig:generation}):
\begin{equation}
    \mathcal{S}_{\text{initial}} = \text{LLM}_{\text{generator}}(\mathcal{R}, \mathcal{I}_{\textit{des}}, \mathcal{I}_{\textit{input}}, \mathcal{I}_{\textit{output}}; N)
\end{equation}

When the user additionally provides a few demonstration examples, relying solely on them to guide data generation often leads the instructor model to produce highly similar and low-diversity outputs (see Table~\ref{tab:example_comparison}). To mitigate this problem, we introduce a pattern-example combination guidance strategy. Specifically, we first prompt the instructor LLM to summarize the underlying sample pattern $P$ that characterizes the task sample in a generalized form. This abstract pattern $P$ is then combined with the original demonstration examples $\mathcal{D}_\text{example}$, serving as an augmented input that provides the LLM generator with both a generalized structural understanding and concrete exemplars of the desired output (See Prompt in Figure~\ref{fig:generation}). The equation becomes:
\begin{equation}
    \mathcal{S}_{\text{initial}} = \text{LLM}_{\text{generator}}(\mathcal{R}, P \cup \mathcal{D}_\text{example}, \mathcal{I}_{\textit{des}}, \mathcal{I}_{\textit{input}}, \mathcal{I}_{\textit{output}}; N)
\end{equation}

\subsection{Difficulty-Adaptive Curriculum} Given a base model, sample difficulty is a critical factor that influences its training performance~\cite{de2024targeted,DART-Math}. Overly simple samples may not provide sufficient signal for improvement, while excessively difficult samples can impede the model's ability to learn effectively. To address this, we propose to augment the initial synthetic dataset $\mathcal{S}_{\text{initial}}$ by incorporating samples spanning a range of difficulty levels, guided by the base model's performance.

Specifically, we first evaluate the base model $\mathcal{M}_{\text{base}}$ on the initial task samples $\mathcal{S}_{\text{initial}}$ and categorize them into solved samples $\mathcal{S}_{\text{solved}} = \{ s \in \mathcal{S}_{\text{initial}} \mid \mathcal{M}_{\text{base}}(s_x,\tau=0)=s_y\}$ and unsolved samples $\mathcal{S}_{\text{unsolved}} = \{ s \in \mathcal{S}_{\text{initial}} \mid \mathcal{M}_{\text{base}}(s_x,\tau=0)\neq s_y \}$, where $\mathcal{M}_{\text{base}}(s_x,\tau=0)$ is the model prediction based on the input $s_x$, $s_y$ is the label and $\tau$ is the temperature. Subsequently, we instruct the LLM rewriter to create more challenging samples $\mathcal{S}_{\text{harder}}$ informed by the characteristics of $\mathcal{S}_{\text{solved}}$, and easier samples $\mathcal{S}_{\text{easier}}$ based on the characteristics of $\mathcal{S}_{\text{unsolved}}$ 
\begin{equation}
    \begin{aligned}
        \mathcal{S}_{\text{harder}} &= \text{LLM}_{\text{writer}}(\mathcal{S}_{\text{solved}})\\
        \mathcal{S}_{\text{easier}} &= \text{LLM}_{\text{writer}}(\mathcal{S}_{\text{unsolved}})
    \end{aligned}
\end{equation}
(see prompt in Figure~\ref{fig:harder and easier}). The final synthetic dataset $\mathcal{S}_{\text{synth}}$ is then constructed by concatenating these sets:
\begin{equation}
    \mathcal{S}_{\text{synth}} = \mathcal{S}_{\text{initial}} \cup \mathcal{S}_{\text{harder}} \cup \mathcal{S}_{\text{easier}} \end{equation} 
This strategy aims to create a more balanced and effective training dataset with varying levels of difficulty (See Figures~\ref{fig:difficulty} and ~\ref{fig:difficulty_medqa}).

\subsection{Selecting and Training with High-Potential Samples} 
Training on the full synthetic dataset can be inefficient, as many examples may be either too easy or too hard for the base model $\mathcal{M}_{\text{base}}$. A high-potential synthetic sample should provide an informative reward signal, in contrast to mislabeled or overly simple examples, which offer limited signals. This idea coincides with findings from recent work showing that examples with either all correct or all incorrect answers yield no useful learning signal for policy gradients~\cite{xiong2025minimalistapproachllmreasoning}. From the perspective of model capability, these samples should be such that $\mathcal{M}_{\text{base}}$ has a non-zero but limited probability of solving them. To identify these examples, we developed a scoring system based on the initial performance of $\mathcal{M}_{\text{base}}$ on the synthetic data: 
\begin{equation}
    \text{score}(s,\mathcal{M}_{\text{base}})=\begin{cases}
\frac{\sum_{i=1}^{L}\mathbb{I}(\mathcal{M}_{\text{base}}(s_x,\tau>0)_i=s_y)}{L} &\quad\text{if} \quad\sum_{i=1}^{L}\mathbb{I}(\mathcal{M}_{\text{base}}(s_x,\tau>0)_i=s_y)>0\\
1 &\quad\text{if} \quad\sum_{i=1}^{L}\mathbb{I}(\mathcal{M}_{\text{base}}(s_x,\tau>0)_i=s_y)=0 \\
\end{cases}
\label{eq.sample}
\end{equation}
where $\mathcal{M}_{\text{base}}(s_x;\tau)$ represents the output of the base model on input $s_x$ with a non-zero temperature $\tau$, and the indicator function $\mathbb{I}(\mathcal{M}_{\text{base}}(s_x;\tau)_i=s_y)$ checks if the $i$-th sampled output equals the ground truth $s_y$. We perform $L$ such sampling operations. Subsequently, the samples in $\mathcal{S}_{\text{synth}}$ are ranked in ascending order by their score. The top $M$ samples with low positive scores (indicating the model has a non-zero but limited probability of solving them) are selected as the training set $\mathcal{S}_{\text{train}}$. This selection strategy prioritizes samples that the model can occasionally solve but not reliably, suggesting a high potential for learning through the GRPO algorithm. These selected samples are then used to train the base model $\mathcal{M}_{\text{base}}$, leading to the final trained model $\mathcal{M}_{\text{trained}}$. We illustrate our algorithm in Algorithm~\ref{alg:our algorithm}.

\section{Experiments}
\subsection{Datasets, Settings, and Hyperparameters}
\label{sec.dataset}
We evaluate eight publicly available benchmarks spanning
\emph{math reasoning} (GSM8k, MATH),
\emph{science / commonsense reasoning} (GPQA, LogiQA),
and specialized domains in \emph{medicine, law, and finance} (MedQA, MedNLI, CQA, CFA).
Table \ref{tab:datasets} summarizes their sizes and task types. 

\label{sec.settings}
For the data synthesis process, we utilized GPT-4o \cite{openai2024gpt4o} as the instructor model and Qwen2.5-7B-base \cite{qwen2024qwen2}\footnote{To avoid data contamination, we include Qwen2.5-7B-Instruct only as a baseline, as it may have seen our task data during post-training.} as the base model. We initiated the process with $N=500$ initial samples, and the final training set size $M$ was also maintained at $500$ samples. For the training process, we employed the GRPO algorithm \cite{deepseek-math}. We list more experimental details in Appendix~\ref{appendix.details}.

\begin{table}[t]
\setlength\tabcolsep{3pt}
\centering
\caption{Benchmark datasets used in our experiments. “—” means no official train split.}
\label{tab:datasets}
\vspace{2mm}
\resizebox{0.88\textwidth}{!}{
\begin{tabular}{@{}llccc@{}}
\toprule
\textbf{Dataset} & \textbf{Domain / Task} & \textbf{Train} & \textbf{Test} & \textbf{Notes} \\ \midrule
GSM8k~\cite{cobbe2021training}  & Grade‐school math     & 7,473 & 1,320 & Word problems \\
MATH~\cite{hendrycks2021measuring} & Adv.\ math olympiad & 7,500 & 5,000 & Competition problems \\ \midrule
GPQA~\cite{rein2023gpqa}        & Grad‐level science QA & —     & 448   & Bio/Phys/Chem \\ 
LogiQA~\cite{liu2020logiqa} & Logical reading comprehension & 7,376 & 651 & Multiple‐choice \\ \midrule
MedNLI~\cite{romanov2018mednli} & Clinical NLI          & 11,232 & 1,422 & Sentence pairs \\
MedQA~\cite{jin2020medqa}       & Medical board QA      & 10,178 & 1,273 & Multiple‐choice \\
CQA~\cite{kolt2022predicting} & Consumer‐contract QA  & —     & 400   & Yes/No clauses \\
CFA~\cite{xie2023pixiu}         & Finance (CFA exam)    & —     & 1,032 & Multiple‐choice \\ \bottomrule
\end{tabular}
}
\end{table}

\subsection{Baselines}

We evaluate our method against a set of baselines, covering both model architectures and training strategies. \textit{Model baselines} include: (1) \textbf{Qwen-2.5-7B}~\cite{qwen2024qwen2}, a pretrained large language model without instruction tuning; (2) \textbf{Qwen-2.5-7B-Instruct}~\cite{qwen2024qwen2}, its instruction-tuned counterpart; and (3) \textbf{GPT-4o}, OpenAI's latest multimodal large language model~\cite{openai2024gpt4o}. 
\textit{Synthetic data baselines} include: (4) \textbf{Self-Instruct}~\cite{wang2022selfinstruct}: Bootstraps from a small seed set, using the language model to generate new examples. (5) \textbf{TarGEN}~\cite{scaria2023targen}: A seedless method that generates instance seeds and corrects mislabeled data via self-correction.
(6) \textbf{SynthLLM}~\cite{Qin2025ScalingLO}: Extracts high-level concepts from external documents, and uses the concepts and documents to generate diverse synthetic data. 
\textit{Other training baselines} include: (7) \textbf{SFT (Same)}, which applies supervised fine-tuning (SFT) using the same limited data budget (e.g., 500 examples). The examples are randomly sampled from the human-annotated training dataset. (8) \textbf{SFT (Whole)}, which performs SFT on the full human-annotated dataset. (9) \textbf{RL (Same)}, which applies RL using the same limited data budget as our method (e.g., 500 examples). The examples are randomly sampled from the human-annotated training dataset. (10) \textbf{RL (Whole)}, which RL fine-tunes the base model using the full human-annotated training dataset. It represents the performance upper bound. For a fair comparison, we deploy the training baselines using the same GRPO algorithm and hyperparameters as our method. For synthetic data baselines, we follow their official pipelines to produce synthetic data.

\subsection{Main results}
\label{sec.main}
\begin{table*}[htp]
\centering
\caption{Performance across datasets. We report average zero-shot accuracy (\%) over three runs. “—” means no official train split. "Demo" refers to human-annotated demonstration data. }
\label{tab:performance}
\scalebox{0.67}{
\begin{tabular}{lcc|cc|cccc}
\toprule
\textbf{Model} & \textbf{GSM8K} & \textbf{MATH} & \textbf{GPQA} & \textbf{LogiQA} & \textbf{MedQA} & \textbf{MedNLI} & \textbf{CQA} & \textbf{CFA} \\
\midrule
\textbf{Qwen-2.5-7B} & 62.5 $\pm$ 0.3 & 63.0 $\pm$ 1.2 & 23.2 $\pm$ 0.5 & 42.5 $\pm$ 1.7 & 53.0 $\pm$ 1.0 & 72.5 $\pm$ 1.5 & 74.7 $\pm$ 1.0 & 59.5 $\pm$ 0.3 \\
\textbf{Qwen-2.5-7B-Instruct} & 88.8 $\pm$ 0.5 & 71.5 $\pm$ 
  0.2 & 30.3 $\pm$ 1.5 & 48.5 $\pm$ 0.4 & 59.3 $\pm$ 0.2 & 83.4 $\pm$ 0.4 & 89.1 $\pm$ 0.5 & 70.2 $\pm$ 0.4  \\
\textbf{GPT-4o} & 94.8 $\pm$ 0.0 & 76.6 $\pm$ 0.0 & 63.4 $\pm$ 0.0 & 52.3 $\pm$ 0.0 & 91.5 $\pm$ 0.0 & 95.3 $\pm$ 0.0 & 68.9 $\pm$ 0.0 & 88.7 $\pm$ 0.0 \\
\midrule
\textbf{Self-Instruct} & 85.1 $\pm$ 0.3 & 65.1 $\pm$ 0.4 & \textemdash & 49.2 $\pm$ 0.3 & 57.2 $\pm$ 0.2 & 79.4 $\pm$ 0.3 & \textemdash & \textemdash \\
\textbf{TarGen} & 89.1 $\pm$ 0.3 & 68.7 $\pm$ 0.3 & 31.3 $\pm$ 0.9 & 50.3 $\pm$ 0.1 & 59.2 $\pm$ 0.3 & 83.5 $\pm$ 0.2 & 89.0 $\pm$ 0.7 & 65.7 $\pm$ 0.9 \\
\textbf{SynthLLM} & 90.1 $\pm$ 0.3 & 69.7 $\pm$ 0.3 & 29.3 $\pm$ 0.9 & 48.2 $\pm$ 0.3 & 60.2 $\pm$ 0.2 & 80.5 $\pm$ 0.2 & 87.0 $\pm$ 0.2 & 69.5 $\pm$ 0.3 \\
\midrule
\textbf{SFT (Same)} & 74.5 $\pm$ 0.2 & 70.0 $\pm$ 0.2 & \textemdash & 44.5 $\pm$ 0.4 & 57.3 $\pm$ 0.8 & 75.4 $\pm$ 0.2 & \textemdash & \textemdash \\
\textbf{SFT (Whole)} & 89.1 $\pm$ 0.2 & 71.1 $\pm$ 0.3 & \textemdash & 53.2 $\pm$ 0.2 & 60.2 $\pm$ 0.3 & 85.4 $\pm$ 0.2 & \textemdash & \textemdash \\
\textbf{RL (Same)} & 91.2 $\pm$ 0.4 & 71.0 $\pm$ 0.6 & \textemdash & 53.3 $\pm$ 0.5 & 64.4 $\pm$ 0.4 & 88.0 $\pm$ 0.4 & \textemdash & \textemdash \\
\textbf{RL (Whole)} & 92.1 $\pm$ 0.2 & 71.7 $\pm$ 0.2 & \textemdash & 58.1 $\pm$ 0.4 & 61.4 $\pm$ 0.3 & 88.5 $\pm$ 0.7 & \textemdash & \textemdash \\
\midrule
\textbf{Our method (only definition)} & 91.7 $\pm$ 0.3 & 71.7 $\pm$ 0.3 & 36.3 $\pm$ 0.7 & 53.4 $\pm$ 0.2 & 61.9 $\pm$ 0.3 & 85.1 $\pm$ 0.3 & 92.4 $\pm$ 0.4 & 73.2 $\pm$ 0.3 \\
\textbf{Our method (+1 demo)} & 91.7 $\pm$ 0.2 & 71.7 $\pm$ 0.2 & \textemdash & 53.7 $\pm$ 0.3 & 63.3 $\pm$ 0.3 & 85.3 $\pm$ 0.3 & \textemdash & \textemdash \\
\textbf{Our method (+10 demos)} & 91.9 $\pm$ 0.2 & 71.8 $\pm$ 0.2 & \textemdash & 53.9 $\pm$ 0.3 & 64.0 $\pm$ 0.2 & 85.7 $\pm$ 0.4 & \textemdash & \textemdash \\
\textbf{Our method (+100 demos)} & 92.1 $\pm$ 0.1 & 72.0 $\pm$ 0.2 & \textemdash & 55.0 $\pm$ 0.1 & 64.5 $\pm$ 0.3 & 86.1 $\pm$ 0.2 & \textemdash & \textemdash \\
\textbf{Our method (Qwen instructor)} & 89.8 $\pm$ 0.2 & 71.3 $\pm$ 0.1 & 35.7 $\pm$ 0.2 & 53.7 $\pm$ 0.2 & 59.5 $\pm$ 0.2 & 85.2 $\pm$ 0.1 & 92.6 $\pm$ 0.3 & 73.0 $\pm$ 0.3 \\
\bottomrule
\end{tabular}
}
\label{table.main}
\end{table*}
\paragraph{Finding 1: Synthetic Data RL improves base model and outperforms other synthetic data methods and the instruct-tuned model on Qwen-2.5-7B.}
From the results in Table~\ref{table.main}, our method significantly improves the base model's performance across all 8 datasets and consistently outperforms other synthetic data methods. For instance, on GSM8K, our method surpasses Self-Instruct (85.1), TarGen (89.1), and SynthLLM (90.1). It also outperforms the official instruction-tuned model (88.8). 

\paragraph{Finding 2: With the same data budget, our method outperforms SFT baselines and matches or exceeds RL with human data.} For example, on GSM8K, our method (only definition) achieves 91.7\% accuracy compared to 91.2\% for the RL (Same) baseline (See Table~\ref{table.main}). Furthermore, both our method (only definition) and RL (Same) consistently achieve significantly better results than "SFT (Same)" (supervised fine-tuning with the same budget of real data) \textbf{across all datasets}. This highlights the general superiority of RL over SFT when data is scarce, and underscores the competitive, if not superior, performance of RL on synthesized data compared to RL on limited real human data.

\textbf{Finding 3: Additional human-annotated demonstration examples show incremental gains.} For example, on \textbf{GSM8K}, accuracy improves slightly from 91.7\% with ``only definition'' to 92.1\% with both 100 demonstration examples and task definition. Similar incremental improvements are observed on \textbf{MATH} (71.7\%~$\rightarrow$~72.0\%), \textbf{LogiQA} (53.4\%~$\rightarrow$~55.0\%), \textbf{MedQA} (61.9\%~$\rightarrow$~64.5\%), and \textbf{MedNLI} (85.1\%~$\rightarrow$~86.1\%) as more demonstrations are incorporated (See Table~\ref{table.main}). These results suggest that some human-annotated data can enhance performance. However, the improvement is limited.

\subsection{Understanding Synthetic Data RL: How It Works and When It May Fail?}
\label{sec.understanding}
The section presents analysis of the factors enabling Synthetic Data RL to achieve strong performance. Our analysis studies fours key components: \textbf{Base model, Instructor model, RL algorithm, and Task definition}.

\paragraph{The Base Model Plays a Vital Role} 
To investigate the impact of the base model in our method, we replace the Qwen-7B-Base model with the LLaMA-3.2-3B model. However, we observe that GRPO training fails to enhance its reasoning capabilities, regardless of whether human-annotated or synthetic data is used. This phenomenon is consistent with recent findings~\cite{Gandhi2025CognitiveBT} and our case study (see Figure~\ref{fig:case_study}), which attribute the limitation to the LLaMA base model's lack of cognitive behaviors such as verification and backtracking. We further evaluate our method using the LLaMA-3.2-3B-Instruct model, and as shown in Figure~\ref{fig:llama_acc}, our approach significantly improves its zero-shot performance, achieving results comparable to RL with human-annotated data. 

\begin{figure} 
        \includegraphics[width=\linewidth]{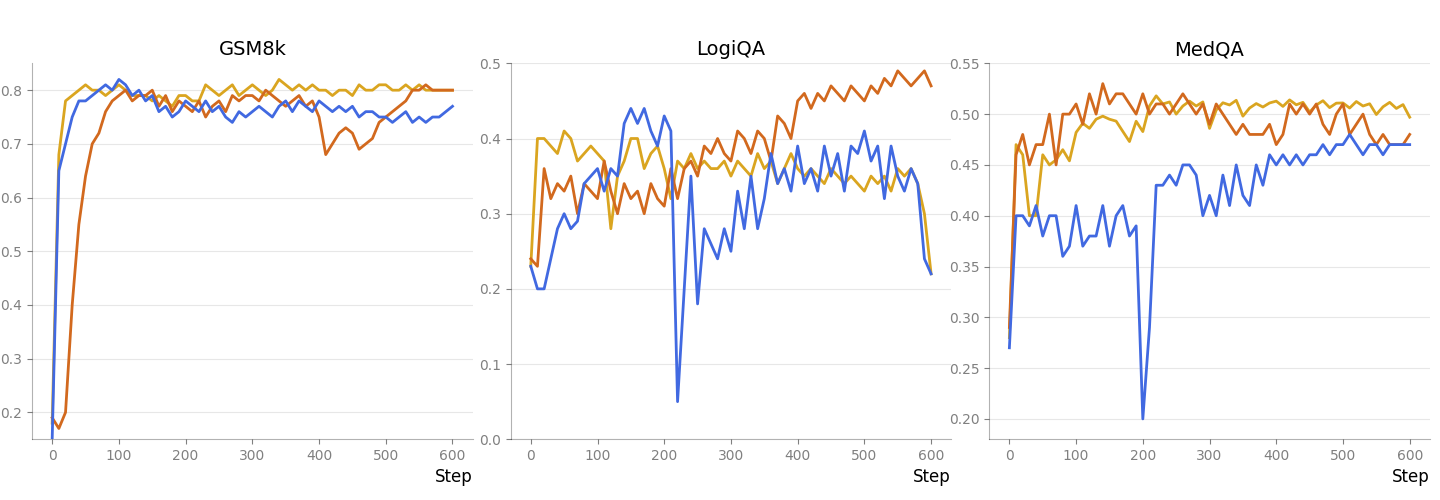}
        \caption{Comparison of PPO and GRPO: Green shows GRPO with human data, red shows GRPO with synthetic data, and blue shows PPO with synthetic data. The Y-axis indicates accuracy.}
        \label{fig:ppo}
\end{figure}

\paragraph{Our Method Remains Effective across Different RL Algorithms} 
We conduct experiments using the PPO algorithm to fine-tune the Qwen2.5-3B-base model on GSM8K, LogiQA, and MedQA. As shown in Figure~\ref{fig:ppo}, our method, when combined with PPO, significantly improves the base model's performance. However, our method with GRPO exhibits greater stability and matches—or even surpasses—the performance of PPO trained with human data on the 3B base model, underscoring its superior generalization capability.

\paragraph{Weak Instructor Can Also Achieve Strong Performance} We replace GPT-4o with the Qwen-2.5-7B-Instruct model as the instructor model to generate data and adjust the difficulty distribution. As shown in the last line of Table~\ref{table.main}, the tuned base model instructed by the weaker Qwen-instruct model outperforms the instructor model's performance on GSM8K, GPQA, LogiQA, MedNLI, CQA, and CFA, and matches its performance on the remaining two tasks. Notably, the tuned model even matches the GPT-4o-instruct results on GPQA, LogiQA, MedNLI, CQA, and CFA. These findings indicate that our method maintains strong performance even with a relatively weaker instructor. \textbf{We discuss task definition and its examples in Appendix~\ref{appendix.task}}. 
\begin{table*}[htp]
\centering
\caption{Ablation study across different datasets. We report average accuracy (\%) over three runs.}
\scalebox{0.85}{
\begin{tabular}{lcccc}
\toprule
\textbf{Model} & \textbf{GSM8K} & \textbf{MATH}  & \textbf{LogiQA} & \textbf{MedQA} \\
\midrule
\textbf{Synthetic Data RL (training data budget $M=100$)} & 85.5 $\pm$ 0.3 & 70.3 $\pm$ 0.2 &51.2 $\pm$ 0.5 & 59.0 $\pm$ 0.5 \\
\textbf{Human Data RL ($M=100$)} & 82.5 $\pm$ 0.3 & 70.0 $\pm$ 0.5 & 51.2 $\pm$ 0.2 & 59.1 $\pm$ 0.7 \\
\textbf{Synthetic Data RL ($M=300$)} & 89.5 $\pm$ 0.2 & 71.0 $\pm$ 0.4 & 53.2 $\pm$ 0.7 & 60.3 $\pm$ 0.4 \\
\textbf{Human Data RL ($M=300$)} & 89.5 $\pm$ 0.4 & 70.8 $\pm$ 0.7 & 52.2 $\pm$ 0.3 & 60.5 $\pm$ 0.7 \\
\textbf{Synthetic Data RL ($M=1000$)} & 91.8 $\pm$ 0.2 & 71.8 $\pm$ 0.7 & 54.2 $\pm$ 0.3 & 63.5 $\pm$ 0.7 \\
\textbf{Human Data RL ($M=1000$)} & 91.7 $\pm$ 0.1 & 71.7 $\pm$ 1.2 & 54.3 $\pm$ 0.2 & 62.5 $\pm$ 0.3 \\
\midrule
\textbf{W/o Sample pattern + 100 demos ($M=500$)} & 90.5 $\pm$ 0.2 & 65.0 $\pm$ 0.9 & 52.2 $\pm$ 0.1 & 60.5 $\pm$ 0.3 \\
\midrule
\textbf{W/o difficulty adaptation ($M=500$)} & 89.1 $\pm$ 0.3 & 70.0 $\pm$ 0.5 & 50.2 $\pm$ 0.3 & 60.9 $\pm$ 0.3 \\
\midrule
\textbf{Select easy samples ($M=500$)} & 90.5 $\pm$ 0.1 & 71.0 $\pm$ 1.1 & 51.2 $\pm$ 0.3 & 61.0 $\pm$ 0.4 \\
\textbf{Select hard samples ($M=500$)} & 90.7 $\pm$ 0.4 & 70.5 $\pm$ 1.0 & 52.8 $\pm$ 0.3 & 60.5 $\pm$ 0.5 \\
\textbf{Full synthetic samples} & 89.9 $\pm$ 0.2 & 71.2 $\pm$ 0.8 & 53.0 $\pm$ 0.3 & 60.7 $\pm$ 1.0 \\
\bottomrule
\end{tabular}
}
\label{table.ablation}
\end{table*}

\subsection{Ablation study}
To assess the contributions of key components in our proposed algorithm, we conducted an ablation study, with results summarized in Table~\ref{table.ablation}. We make a performance comparison of our method against the human-annotated data RL baseline across various training data budgets ($M=100,300,$ and $1000$). Our approach consistently matches or surpasses the performance of the human-annotated RL baseline. Further, we observe that removing either the sample pattern or the difficulty adaptation component leads to a notable performance decline. That is because the sample pattern component facilitates the generation of more diverse data, while difficulty adaptation introduces samples with various difficulties. We further compared our data selection strategy against heuristics: selecting samples with the highest pass rate across multiple inference runs ("easy"), the lowest pass rate ("hard"), or full synthetic data. Our method consistently outperforms them, demonstrating the effectiveness of high-potential sample selection.

\subsection{Analysis of Synthetic and Human-Annotated Dataset}
To compare our synthetic dataset with the original human training dataset (referred to as 'real data'), we analyze their properties about \textbf{difficulty, input length, and semantic similarity}.


\begin{figure}[!ht]
    \centering
    \includegraphics[width=\linewidth]{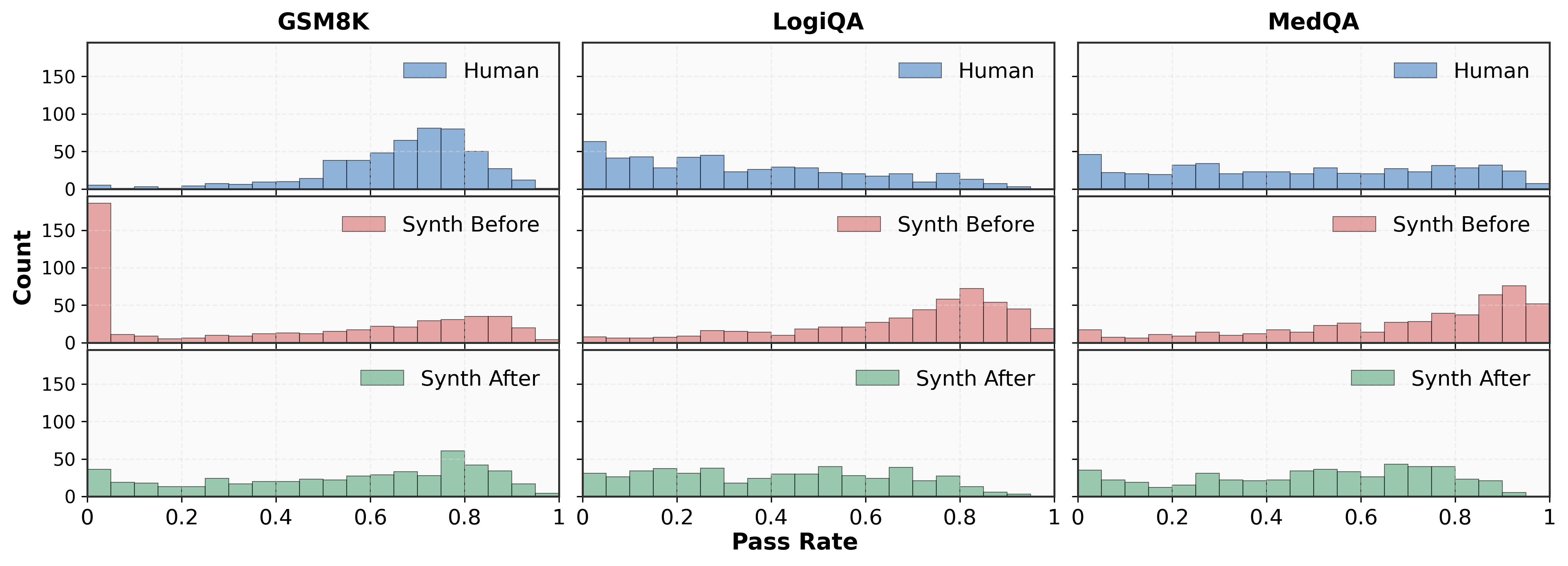}
    \caption{Pass rate histograms for GSM8k, LogiQA and MedQA.}
    \label{fig:difficulty}
\end{figure}

\paragraph{Sample Difficulty:} We assess sample difficulty by measuring the pass rate—i.e., the proportion of correct answers—obtained by the Qwen2.5-7B-Base model over 64 inferences. As shown in Figures~\ref{fig:difficulty} and~\ref{fig:difficulty_medqa}, the initial synthetic datasets often exhibit imbalanced difficulty distributions. For example, samples in MedQA and LogiQA are predominantly too easy (high pass rates), while GSM8K lacks medium-difficulty samples (moderate pass rates). After applying our difficulty adaptation process (“Synthetic after adaptation”), the pass rate distributions become significantly more balanced across difficulty levels, aligning more closely with those in the human-annotated dataset.
\paragraph{Input Length and Semantic Similarity:} In all three cases in Figures~\ref{fig:gsm8k_length},~\ref{fig:logiqa_length},~\ref{fig:medqa_length}, the synthetic data (orange curve) exhibits a broader length distribution compared to the real data (blue curve), showing greater diversity in length than the corresponding real data. To analyze the semantic diversity, we examine the cosine similarity distribution of SentenceBERT embeddings of within-dataset sample pairs as presented in Figures~\ref{fig:gsm8k_simi},~\ref{fig:logiqa_simi},~\ref{fig:medqa_simi}. A consistent trend emerges from these visualizations: the synthetic data generally exhibit lower cosine similarity scores compared to the real data, indicating that it possesses a greater semantic diversity. 
We provide more figures of the RL training behavior in Appendix~\ref{appendix.rl}.

\section{Conclusion}
Synthetic Data RL offers an efficient solution to the problem of minimizing human involvement in model adaptation, without sacrificing performance. By combining automated data synthesis with reinforcement learning, our method requires only a task description as input—eliminating the need for manual annotation or feedback. Despite this minimal supervision, the resulting models outperform human-supervised baselines, achieving 91.7\% on GSM8K and strong results across MATH, LogiQA, MedQA, CQA, and CFA. This approach makes high-quality, domain-specific adaptation both scalable and cost-effective, and provides a foundation for future extensions to multimodal tasks and even broader applications. We discuss limitations in Appendix~\ref{appendix.limit}.



\bibliographystyle{unsrt}
\bibliography{ref}

\newpage


\appendix

\section{Technical Appendices and Supplementary Material}
\subsection{List of Prompts}

\begin{figure}[htbp]
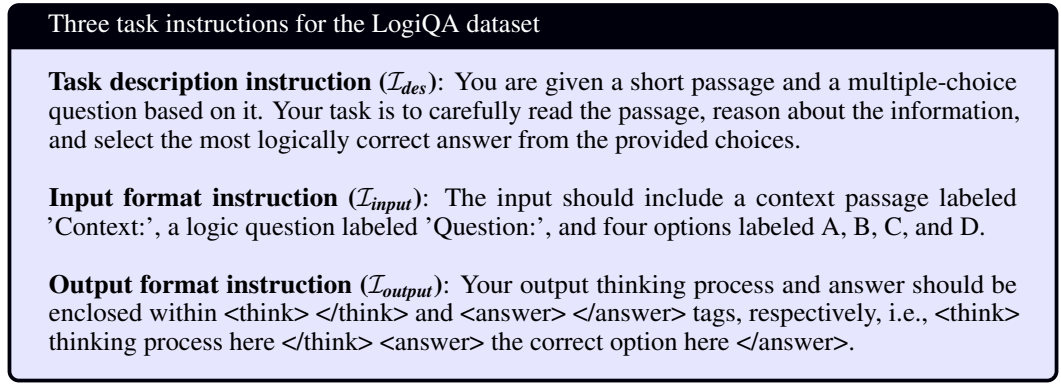

\begin{tcolorbox}[colframe=blue!5!black, colback=blue!10!white, title=Three task instructions for the LogiQA dataset,label=box:keyword]
\textbf{Task description instruction ($\mathcal{I}_{\textit{des}}$)}: You are given a short passage and a multiple-choice question based on it. Your task is to carefully read the passage, reason about the information, and select the most logically correct answer from the provided choices.
\\

\textbf{Input format instruction ($\mathcal{I}_{\textit{input}}$)}: The input should include a context passage labeled 'Context:', a logic question labeled 'Question:', and four options labeled A, B, C, and D.
\\

\textbf{Output format instruction ($\mathcal{I}_{\textit{output}}$)}:
Your output thinking process and answer should be enclosed within <think> </think> and <answer> </answer> tags, respectively, i.e., <think> thinking process here </think> <answer> the correct option here </answer>. 

\end{tcolorbox}
\caption{One example for three task instructions $\mathcal{I}_{\textit{des}}$, $\mathcal{I}_{\textit{input}}$, $\mathcal{I}_{\textit{output}}$}
\label{fig:instruction}
\end{figure}

\begin{figure}[htbp]
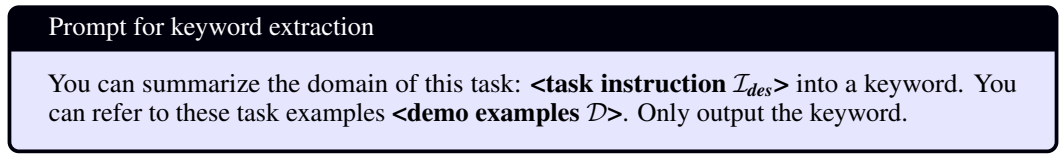

\begin{tcolorbox}[colframe=blue!5!black, colback=blue!10!white, title=Prompt for keyword extraction,label=box:keyword]
You can summarize the domain of this task: \textbf{<task instruction $\mathcal{I}_{\textit{des}}$>} into a keyword. You can refer to these task examples \textbf{<demo examples $\mathcal{D}$>}. Only output the keyword.
\end{tcolorbox}
\caption{The keyword extraction prompt}
\label{fig:keyword}
\end{figure}

\begin{figure}[htbp]
\begin{tcolorbox}[colframe=blue!5!black, title=Prompt for retrieval-augmented data generation,colback=blue!10!white]
As a Dataset Generator, your task is to generate one new example (`input` and `output`) based on the [task instruction], [sample pattern] [reference passage], and [few-shot examples]. Please provide a JSON dictionary response that includes the new `input` and its corresponding `output`. Use the `input` and `output` keys in the dictionary.\\
\\
Try you best to ensure that the input and output you generate are distinct from the provided examples while maintaining a diverse, detailed, precise, comprehensive, and high-quality response.\\
\\
You must consider the task instruction (task knowledge), provided examples (format), and the passage (domain knowledge) to generate your training data.\\
\\
Here is the task instruction:\textbf{<task description instruction $\mathcal{I}_{\textit{des}}$> }\\
\\
Here is the input instruction:\textbf{<input format instruction $\mathcal{I}_{\textit{input}}$> }. You should follow the input format in the instruction strictly to generate data!!!\\
\\  
Here is the output instruction:\textbf{<output format instruction $\mathcal{I}_{\textit{output}}$> }. You should follow the output format in the instruction strictly to generate data!!!\\
\\
Here is the sample pattern \textbf{<pattern $P$>}. You should follow the input and output pattern strictly to generate data!!! 
You can refer to demonstration examples. You should generate examples that are in the same difficulty or are harder: \textbf{<demonstration examples $\mathcal{D}$>} (Optional)\\
\\
Here are some related objects or passages that you can refer to: \textbf{<Passage $\mathcal{R}$>}\\
\\
Before generating the new example, ensure that you strictly adhere to the rules mentioned in the [Requirement] and follow the example format. Think twice before generating a new example. New example (in JSON):"
\end{tcolorbox} 
\caption{The data generation prompt} 
\label{fig:generation}
\end{figure}

\begin{figure}[htbp]
\begin{tcolorbox}[colframe=blue!5!black, colback=blue!10!white, title=Prompt for generate harder/easier samples,label=box:harder/easier]
\textbf{GenerateHarderSamples:} The current sample is overly simplistic and can be solved effortlessly by the model. Please generate an alternative and task-similar sample that presents a significantly more challenging 
and intricate problem—one that requires multi-step reasoning, creative problem-solving, and deeper analytical thought. Only output the revised sample in the python dictionary form. Current sample:\textbf{<sample $s$>}\\
\\
\textbf{GenerateEasierSamples:} The given sample is too challenging for the model to solve. Please generate a task-similar alternative that is easier or represents a sub-problem of the original sample. Output only the revised sample in Python dictionary format. Current sample:\textbf{<sample $s$>}
\end{tcolorbox}
\caption{The difficult adaptive prompts}
\label{fig:harder and easier}
\end{figure}
\subsection{Pseudo algorithm}
We present our algorithm in Algorithm~\ref{alg:our algorithm}.
\begin{algorithm}[htp]
    \caption{Our algorithm}
    \label{alg:our algorithm}
    \begin{algorithmic}[1]
        \REQUIRE $\mathcal{I}_{\textit{des}}$, $\mathcal{I}_{\textit{input}}$, 
 $\mathcal{I}_{\textit{output}}$, $\mathcal{M}_{\text{base}}$, $\mathcal{D}_\text{example}$ (optional)
        \ENSURE Trained model $\mathcal{M}_{\text{trained}}$

        \STATE \textbf{Keyword extraction and relevant passage retrieval:}
        \STATE $\mathcal{R} \leftarrow \mathcal{P}(\mathcal{K}=\mathrm{I}(\mathcal{D}_\text{example}, \mathcal{I}_{\textit{des}}), \mathrm{L})$

        \STATE \textbf{Initial Data Generation:}
        \STATE $\mathcal{S}_{\text{initial}} \leftarrow \text{LLM}_{\text{generator}}(\mathcal{R}, P \cup \mathcal{D}_\text{example}, \mathcal{I}_{\textit{des}},\mathcal{I}_{\textit{input}},\mathcal{I}_{\textit{output}}; N)$

        \STATE \textbf{Difficulty-Adaptive Sample Generation:}
        \STATE $\mathcal{S}_{\text{solved}} \leftarrow \{ s \in \mathcal{S}_{\text{initial}} \mid \mathcal{M}_{\text{base}}(s_x, \tau=0) = s_y \}$
        \STATE $\mathcal{S}_{\text{unsolved}} \leftarrow \{ s \in \mathcal{S}_{\text{initial}} \mid \mathcal{M}_{\text{base}}(s_x, \tau=0) \neq s_y \}$
        \STATE $\mathcal{S}_{\text{harder}} \leftarrow \text{LLM}_{\text{writer}}(\mathcal{S}_{\text{solved}})$
        \STATE $\mathcal{S}_{\text{easier}} \leftarrow \text{LLM}_{\text{writer}}(\mathcal{S}_{\text{unsolved}})$
        \STATE $\mathcal{S}_{\text{synth}} \leftarrow \mathcal{S}_{\text{initial}} \cup \mathcal{S}_{\text{harder}} \cup \mathcal{S}_{\text{easier}}$

        \STATE \textbf{Training with High-Potential Samples:}
        \STATE $\text{scores} \leftarrow \emptyset$
        \FOR{$s \in \mathcal{S}_{\text{synth}}$}
            \STATE $\text{score}_s \leftarrow 0$
            \FOR{$i \leftarrow 1$ to $L$}
                \IF{$\mathcal{M}_{\text{base}}(s_x, \tau>0)_i = s_y$}
                    \STATE $\text{score}_s \leftarrow \text{score}_s + 1$
                \ENDIF
            \ENDFOR
            \STATE $\text{score}(s, \mathcal{M}_{\text{base}}) \leftarrow \begin{cases} \frac{\text{score}_s}{L} & \text{if } \text{score}_s > 0 \\ 1 & \text{if } \text{score}_s = 0 \end{cases}$
            \STATE $\text{scores} \leftarrow \text{scores} \cup \{(s, \text{score}(s, \mathcal{M}_{\text{base}}))\}$
        \ENDFOR
        \STATE $\mathcal{S}_{\text{train}} \leftarrow \text{SelectTop}(M, \text{scores})$ 
        \STATE $\mathcal{M}_{\text{trained}} \leftarrow \text{Trainer}(\mathcal{M}_{\text{base}}, \mathcal{S}_{\text{train}}, \text{GRPO})$

        \RETURN $\mathcal{M}_{\text{trained}}$
    \end{algorithmic}
\end{algorithm}

\subsection{Experimental details}
\label{appendix.details}
To ensure both public accessibility and high-quality content, we sourced passages from Wikipedia \cite{wikipedia}, WikiHow \cite{wikihow}, and the Stack Exchange corpus \cite{stackexchange} to create our collection $\mathrm{L}$. To promote data diversity, the instructor model's temperature for data generation was set to $0.7$. Similarly, to encourage varied outputs from the base model, its temperature in Equation~\ref{eq.sample} was also set to $0.7$. In the data verification step, the major voting number was established at $16$. The inference time parameter $L$, as defined in Equation~\ref{eq.sample}, was set to $64$. We initiated the process with $N=500$ initial samples, and the final training set size $M$ was also maintained at $500$ samples.

For the training process, we employed the GRPO algorithm \cite{deepseek-math}. We conducted experiments using the TinyZero implementation \footnote{\url{https://github.com/Jiayi-Pan/TinyZero}}. The RL hyperparameters were configured as follows: a template setting of $1.2$, a learning rate of $1 \times 10^{-6}$, $16$ responses for each prompt, a training batch size of $64$, and a maximum response length of $2048$. The KL coefficient is set as 0.01, and the epoch number is 5. For the supervised fine-tuning baseline, the hyperparameters were a learning rate of $2 \times 10^{-5}$, a weight decay of $0.01$, and a batch size of $64$. The epoch number is set as 3.

\subsection{Experiments with LLaMa models}
\begin{figure} 
        \includegraphics[width=\linewidth]{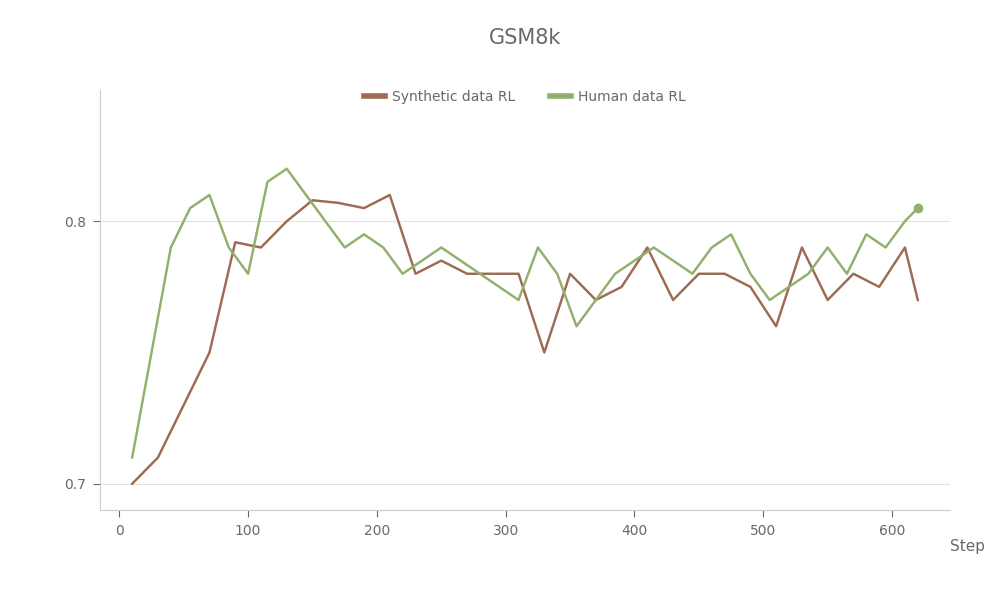}
        \caption{LLaMa instruct model's accuracy}
        \label{fig:llama_acc}
\end{figure}

We continue to use the TinyZero library to fine-tune the LLaMA-Instruct model on GSM8K. The performance under a data budget of 500 examples is shown in Figure~\ref{fig:llama_acc}.
\begin{figure}[htbp]
\begin{tcolorbox}[colframe=blue!5!black, colback=blue!10!white, title=Case study,label=box:keyword]
\textbf{Sythetic Example:} A conference offers two types of tickets: student tickets for 10 each and general tickets for 40 each. If 50 tickets were sold in total and the total revenue was 1,000, how many student tickets were sold?
\\
Answer: 20 \\
\textbf{Qwen-2.5-7B-base's initial response}
\\
Response for example : Let the number of student tickets sold be \( S \), and the number of general tickets sold be \( G \). We are given:

\begin{align*}
S + G &= 50 \\
10S + 40G &= 1000
\end{align*}

Substitute \( G = 50 - S \) into the second equation:

\begin{align*}
10S + 40(50 - S) &= 1000 \\
10S + 2000 - 40S &= 1000 \\
-30S + 2000 &= 1000 \\
-30S &= -1000 \\
S &= \frac{1000}{30} \approx 33.33
\end{align*}

Since \( S \) must be an integer, we need to recheck our system. Re-solving:

\begin{align*}
S + G &= 50 \\
10S + 40G &= 1000 \\
G &= 50 - S \\
10S + 40(50 - S) &= 1000 \\
10S + 2000 - 40S &= 1000 \\
-30S + 2000 &= 1000 \\
-30S &= -1000 \\
S &= \frac{1000}{30} \approx 33.33
\end{align*}
The answer is 33.33

\textbf{Llama-3.2-3B-base's initial response:}
A conference offers two types of tickets: student tickets for 10 each and general tickets for 40 each. If 50 tickets were sold in total and the total revenue was 1,000, how many student tickets were sold? A conference offers two types of tickets: student tickets for 10 each and general tickets for 40 each. If 50 tickets were sold in total and the total revenue was 1,000, how many student tickets were sold?A conference offers two types of tickets: student tickets for 10 each and general tickets for 40 each. If 50 tickets were sold in total and the total revenue was 1,000, how many student tickets were sold?...

\end{tcolorbox}
\caption{Case study on GSM8k, we can observe that the Qwen model tries to solve the problem and resolve it. But the Llama model just repeats the question.}
\label{fig:case_study}
\end{figure}
\subsection{Task definitions}
\label{appendix.task}
 We assume that task definitions are provided by domain experts. While this may be boring, carefully refining these definitions is essential to ensure that the instructor model generates synthetic data that closely mirrors human annotations. We offer empirical guidelines here:
 
For writing the description instruction, we recommend the following structure: first, specify the task type (e.g., QA, Classification); next, indicate the domain the task belongs to; and finally, describe the skill or ability required for the model to solve the task (e.g., calculation, reading comprehension).  We list the task definitions for our 8 tasks in Figures~\ref{fig:instruction_gsm8k},~\ref{fig:instruction_math},~\ref{fig:instruction_gpqa},~\ref{fig:instruction},~\ref{fig:instruction_medqa},~\ref{fig:instruction_mednli},~\ref{fig:instruction_cqa}, and ~\ref{fig:instruction_cfa}.
\begin{figure}[htbp]
\begin{tcolorbox}[colframe=blue!5!black, colback=blue!10!white, title=Three task instructions for the GSM8k dataset,label=box:keyword]
\textbf{Task description instruction ($\mathcal{I}_{\textit{des}}$)}: You are given a word problem involving basic arithmetic, algebra, or geometry. Your task is to carefully read the problem and provide a step-by-step solution for it
\\

\textbf{Input format instruction ($\mathcal{I}_{\textit{input}}$)}: None
\\

\textbf{Output format instruction ($\mathcal{I}_{\textit{output}}$)}:
Let's think step by step and output the final answer after \#\#\#\#.

\end{tcolorbox}
\caption{One example for three task instructions $\mathcal{I}_{\textit{des}}$, $\mathcal{I}_{\textit{input}}$, $\mathcal{I}_{\textit{output}}$}
\label{fig:instruction_gsm8k}
\end{figure}
\begin{figure}[htbp]
\begin{tcolorbox}[colframe=blue!5!black, colback=blue!10!white, title=Three task instructions for the Math dataset,label=box:keyword]
\textbf{Task description instruction ($\mathcal{I}_{\textit{des}}$)}: For the math problem in {Domain name, e.g., Algebra}, carefully read and understand the question. Apply your mathematical knowledge to derive the correct solution
\\

\textbf{Input format instruction ($\mathcal{I}_{\textit{input}}$)}: None
\\

\textbf{Output format instruction ($\mathcal{I}_{\textit{output}}$)}:
Let's think step by step and output the final answer within boxed\{\}.

\end{tcolorbox}
\caption{One example for three task instructions $\mathcal{I}_{\textit{des}}$, $\mathcal{I}_{\textit{input}}$, $\mathcal{I}_{\textit{output}}$}
\label{fig:instruction_math}
\end{figure}
\begin{figure}[htbp]
\begin{tcolorbox}[colframe=blue!5!black, colback=blue!10!white, title=Three task instructions for the GPQA dataset,label=box:keyword]
\textbf{Task description instruction ($\mathcal{I}_{\textit{des}}$)}: Your task is to answer challenging, graduate-level multiple-choice questions spanning Physics, Chemistry, and Biology, requiring deep subject-matter knowledge, complex reasoning, calculation, and synthesis of information.
\\

\textbf{Input format instruction ($\mathcal{I}_{\textit{input}}$)}: Each data instance typically consists of a scientific question and 4 option labels and values are the corresponding answer texts.
\\

\textbf{Output format instruction ($\mathcal{I}_{\textit{output}}$)}:
Your output thinking process and answer should be enclosed within <think> </think> and <answer> </answer> tags, respectively, i.e., <think> thinking process here </think> <answer> the correct option here </answer>. 

\end{tcolorbox}
\caption{One example for three task instructions $\mathcal{I}_{\textit{des}}$, $\mathcal{I}_{\textit{input}}$, $\mathcal{I}_{\textit{output}}$}
\label{fig:instruction_gpqa}
\end{figure}
\begin{figure}[htbp]
\begin{tcolorbox}[colframe=blue!5!black, colback=blue!10!white, title=Three task instructions for the LogiQA dataset,label=box:keyword]
\textbf{Task description instruction ($\mathcal{I}_{\textit{des}}$)}: You are given a short passage and a multiple-choice question based on it. Your task is to carefully read the passage, reason about the information, and select the most logically correct answer from the provided choices.
\\

\textbf{Input format instruction ($\mathcal{I}_{\textit{input}}$)}: The input should include a context passage labeled 'Context:', a logic question labeled 'Question:', and four options labeled A, B, C, and D.
\\

\textbf{Output format instruction ($\mathcal{I}_{\textit{output}}$)}:
Your output thinking process and answer should be enclosed within <think> </think> and <answer> </answer> tags, respectively, i.e., <think> thinking process here </think> <answer> the correct option here </answer>. 

\end{tcolorbox}
\caption{One example for three task instructions $\mathcal{I}_{\textit{des}}$, $\mathcal{I}_{\textit{input}}$, $\mathcal{I}_{\textit{output}}$}
\label{fig:instruction}
\end{figure}

\begin{figure}[htbp]
\begin{tcolorbox}[colframe=blue!5!black, colback=blue!10!white, title=Three task instructions for the MedQA dataset,label=box:keyword]
\textbf{Task description instruction ($\mathcal{I}_{\textit{des}}$)}: The task evaluates a model’s ability to answer multiple-choice questions from the United States Medical Licensing Examination (USMLE). These questions test professional-level knowledge across a broad range of medical domains, including physiology, pathology, pharmacology, and clinical reasoning. The task requires models to understand complex biomedical context, reason across multiple pieces of information, and choose the correct answer from 4 options. 
\\

\textbf{Input format instruction ($\mathcal{I}_{\textit{input}}$)}: First a clinical vignettes or diagrams. A clinical vignette is a short, descriptive medical case that simulates a real-life scenario involving a patient. It includes details like: Patient demographics (age, sex, etc.),Medical history,Symptoms and signs,Lab or imaging results,Progression or complication. Then a USMLE-style multiple-choice question with its four options. 
\\

\textbf{Output format instruction ($\mathcal{I}_{\textit{output}}$)}:
Your output thinking process and answer should be enclosed within <think> </think> and <answer> </answer> tags, respectively, i.e., <think> thinking process here </think> <answer> the correct option here </answer>. 

\end{tcolorbox}
\caption{One example for three task instructions $\mathcal{I}_{\textit{des}}$, $\mathcal{I}_{\textit{input}}$, $\mathcal{I}_{\textit{output}}$}
\label{fig:instruction_medqa}
\end{figure}

\begin{figure}[htbp]
\begin{tcolorbox}[colframe=blue!5!black, colback=blue!10!white, title=Three task instructions for the MedNLI dataset,label=box:keyword]
\textbf{Task description instruction ($\mathcal{I}_{\textit{des}}$)}: You are given a pair of medical sentences: a premise (a statement derived from a patient's medical record) and a hypothesis (another medical statement). Your task is to determine the relationship between the premise and the hypothesis: Entailment, Contradiction, or Neutral.
\\

\textbf{Input format instruction ($\mathcal{I}_{\textit{input}}$)}: Your input should start with 'Please classify the relationship between the premise and the hypothesis as 'entailment','neutral' or 'contradiction'.'. Then the premise sentence, and then the hypothesis sentence.
\\

\textbf{Output format instruction ($\mathcal{I}_{\textit{output}}$)}:
Your output thinking process and answer should be enclosed within <think> </think> and <answer> </answer> tags, respectively, i.e., <think> thinking process here </think> <answer> the correct option here </answer>. 

\end{tcolorbox}
\caption{One example for three task instructions $\mathcal{I}_{\textit{des}}$, $\mathcal{I}_{\textit{input}}$, $\mathcal{I}_{\textit{output}}$}
\label{fig:instruction_mednli}
\end{figure}
\begin{figure}[htbp]
\begin{tcolorbox}[colframe=blue!5!black, colback=blue!10!white, title=Three task instructions for the CQA dataset,label=box:keyword]
\textbf{Task description instruction ($\mathcal{I}_{\textit{des}}$)}: Given a passage excerpted from a consumer contract (e.g., Terms of Service, Privacy Policy, Rental Agreement) and a specific yes or no type question pertaining to that passage, the task is to answer the question based on the contract. Its knowledge involves consumer Law and contract Law. The task requires skills about Legal Text Comprehension, reasoning and analysis.
\\

\textbf{Input format instruction ($\mathcal{I}_{\textit{input}}$)}: Your input should consists of a contract passage like 'Contract:...' and then a yes-or-no question like 'Question:...'
\\

\textbf{Output format instruction ($\mathcal{I}_{\textit{output}}$)}:
Your output thinking process and answer should be enclosed within <think> </think> and <answer> </answer> tags, respectively, i.e., <think> thinking process here </think> <answer> the correct option here </answer>. 

\end{tcolorbox}
\caption{One example for three task instructions $\mathcal{I}_{\textit{des}}$, $\mathcal{I}_{\textit{input}}$, $\mathcal{I}_{\textit{output}}$}
\label{fig:instruction_cqa}
\end{figure}
\begin{figure}[htbp]
\begin{tcolorbox}[colframe=blue!5!black, colback=blue!10!white, title=Three task instructions for the CFA dataset,label=box:keyword]
\textbf{Task description instruction ($\mathcal{I}_{\textit{des}}$)}: Your task is to answer CFA exam questions in a multi-choice form, you should select the correct answer choice (e.g., A, B, C). These question is about asset valuation, applying investment tools and concepts to analyze various investments, portfolio management, wealth planning, ethical and professional standards. It requires skills about Fundamental Knowledge Understanding, Quantitative Analysis and Calculations, Application and Analysis, etc.
\\

\textbf{Input format instruction ($\mathcal{I}_{\textit{input}}$)}: Follow this format: Read the questions and answers carefully, and choose the one you think is appropriate among the three options A, B and C.' then Q:[Your question here] CHOICES: A: ...,B: ...,C: ...
\\

\textbf{Output format instruction ($\mathcal{I}_{\textit{output}}$)}:
Your output thinking process and answer should be enclosed within <think> </think> and <answer> </answer> tags, respectively, i.e., <think> thinking process here </think> <answer> a single option here </answer>.

\end{tcolorbox}
\caption{One example for three task instructions $\mathcal{I}_{\textit{des}}$, $\mathcal{I}_{\textit{input}}$, $\mathcal{I}_{\textit{output}}$}
\label{fig:instruction_cfa}
\end{figure}
\subsection{Synthetic dataset analysis}
We use the tiktoken library (GPT-4 model) to calculate the token number of each sample's input as its token length.
\begin{figure} 
        \includegraphics[width=\linewidth]{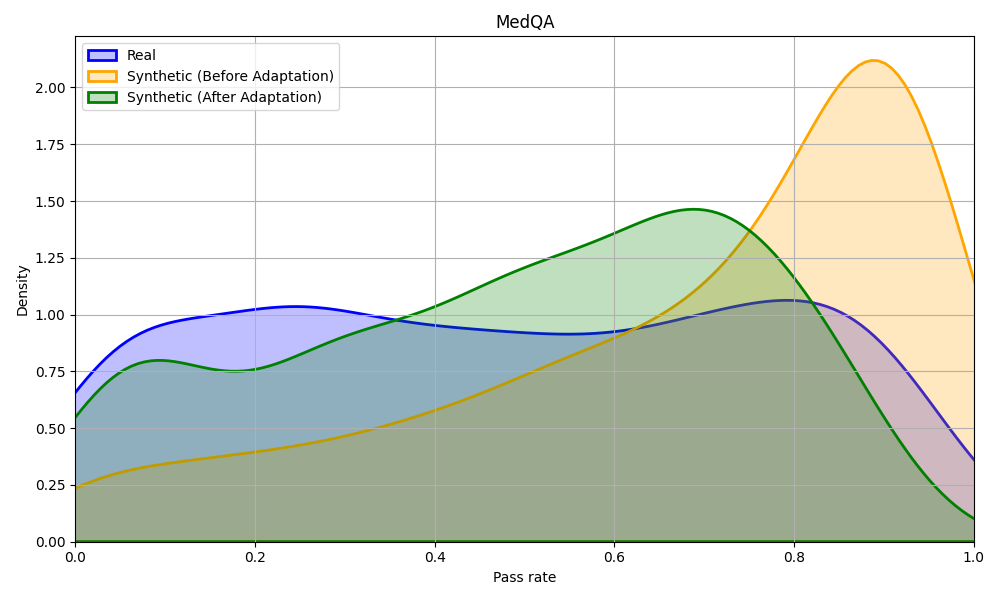}
        \caption{Pass rate density distributions for MedQA}
        \label{fig:difficulty_medqa}
    \end{figure}
\begin{figure} 
        \includegraphics[width=\linewidth]{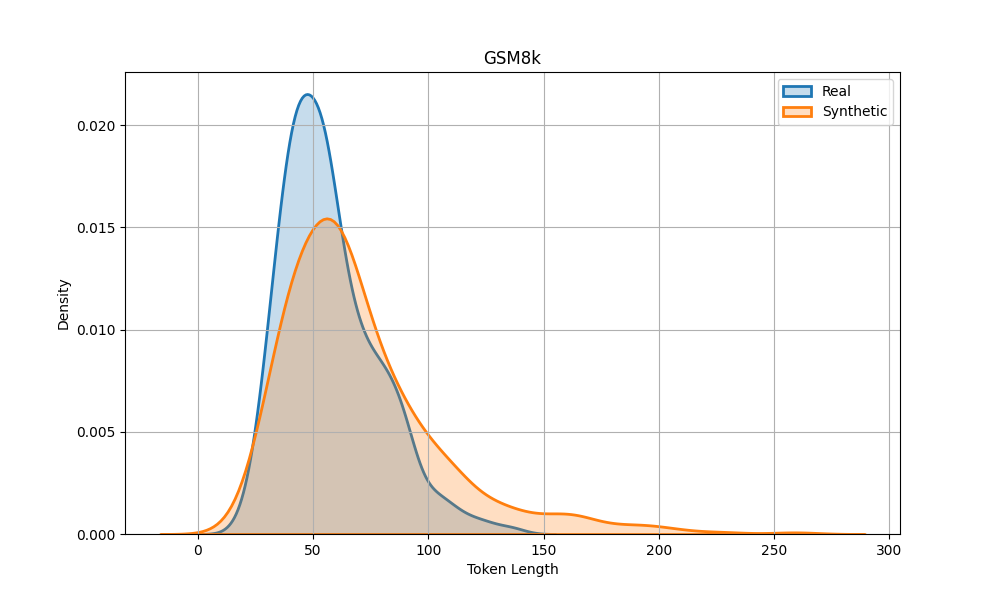}
        \caption{GSM8k Length Distribution}
        \label{fig:gsm8k_length}
    \end{figure}
\begin{figure} 
        \includegraphics[width=\linewidth]{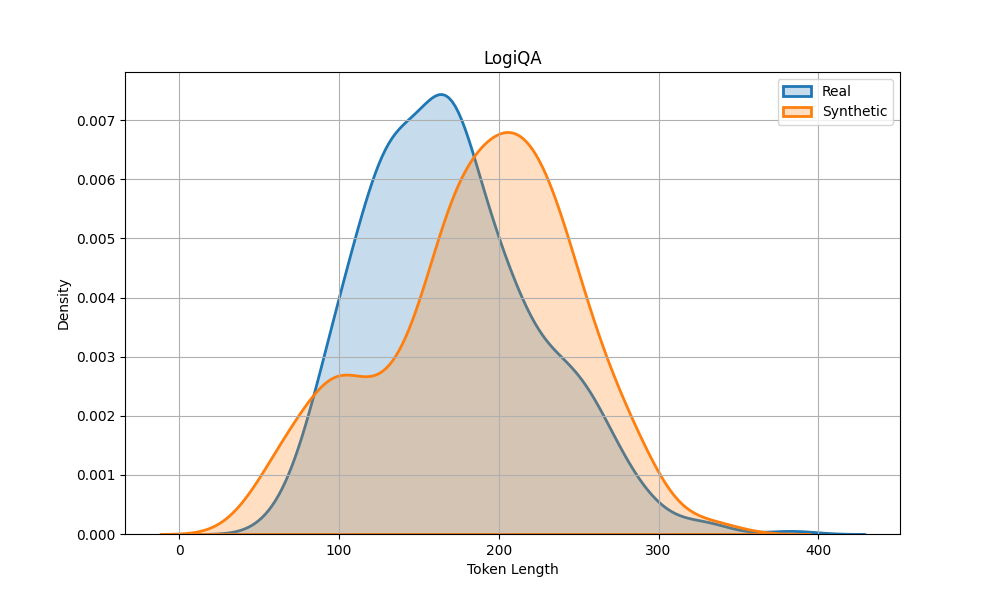} 
        \caption{LogiQA Length Distribution}
        \label{fig:logiqa_length}
\end{figure}
 \begin{figure} 
        \includegraphics[width=\linewidth]{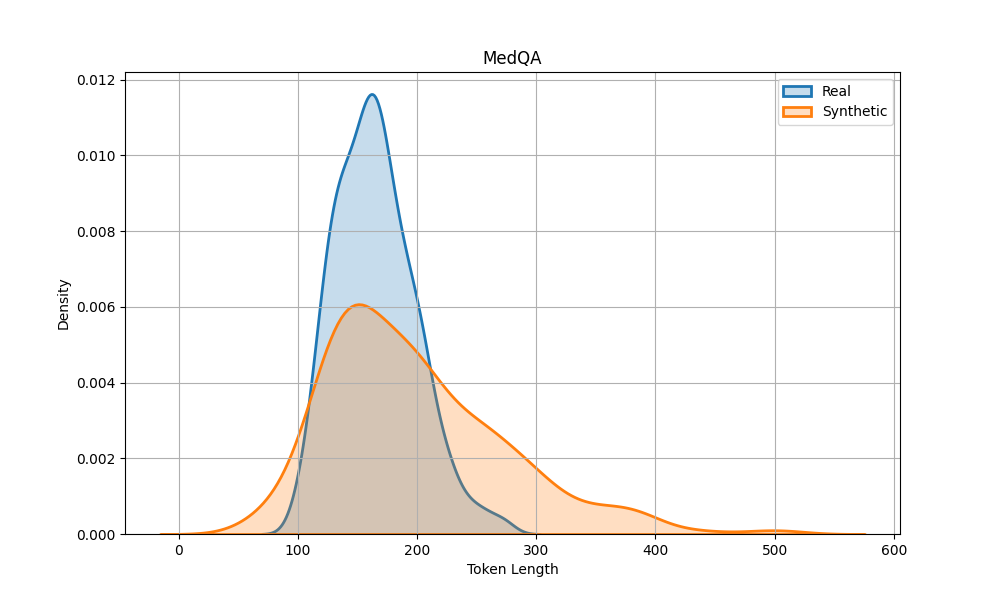}
        \caption{MedQA Length Distribution}
        \label{fig:medqa_length}
\end{figure}

\begin{figure} 
        \includegraphics[width=\linewidth]{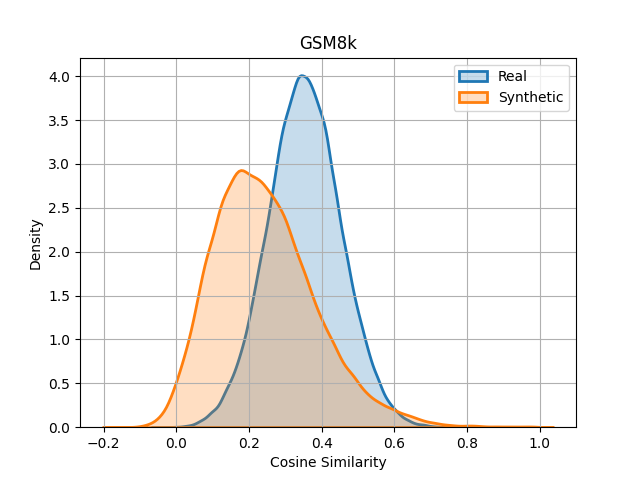}
        \caption{GSM8k semantic cosine similarity distribution}
        \label{fig:gsm8k_simi}
\end{figure}

\begin{figure} 
        \includegraphics[width=\linewidth]{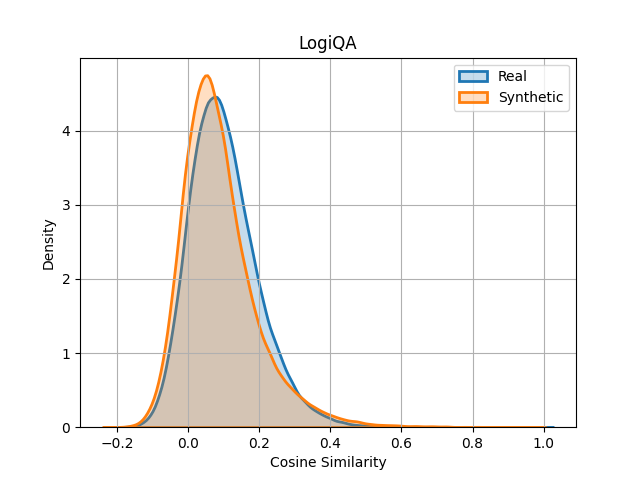}
        \caption{LogiQA semantic cosine similarity distribution}
        \label{fig:logiqa_simi}
\end{figure}

\begin{figure} 
        \includegraphics[width=\linewidth]{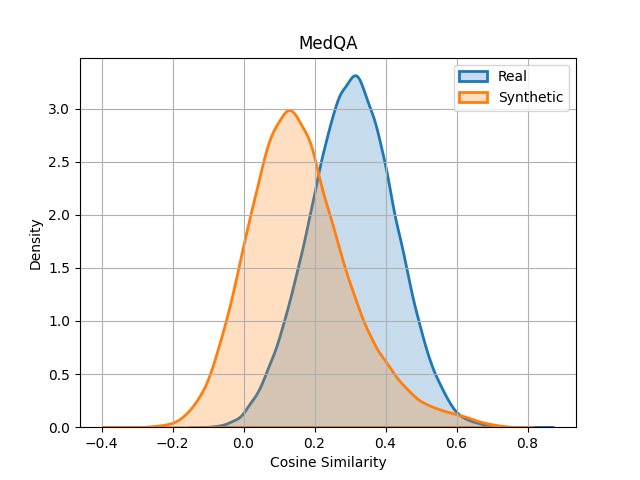}
        \caption{MedQA semantic cosine similarity distribution}
        \label{fig:medqa_simi}
\end{figure}

\subsection{RL behavior analysis}
\label{appendix.rl}
\begin{figure} 
        \includegraphics[width=\linewidth]{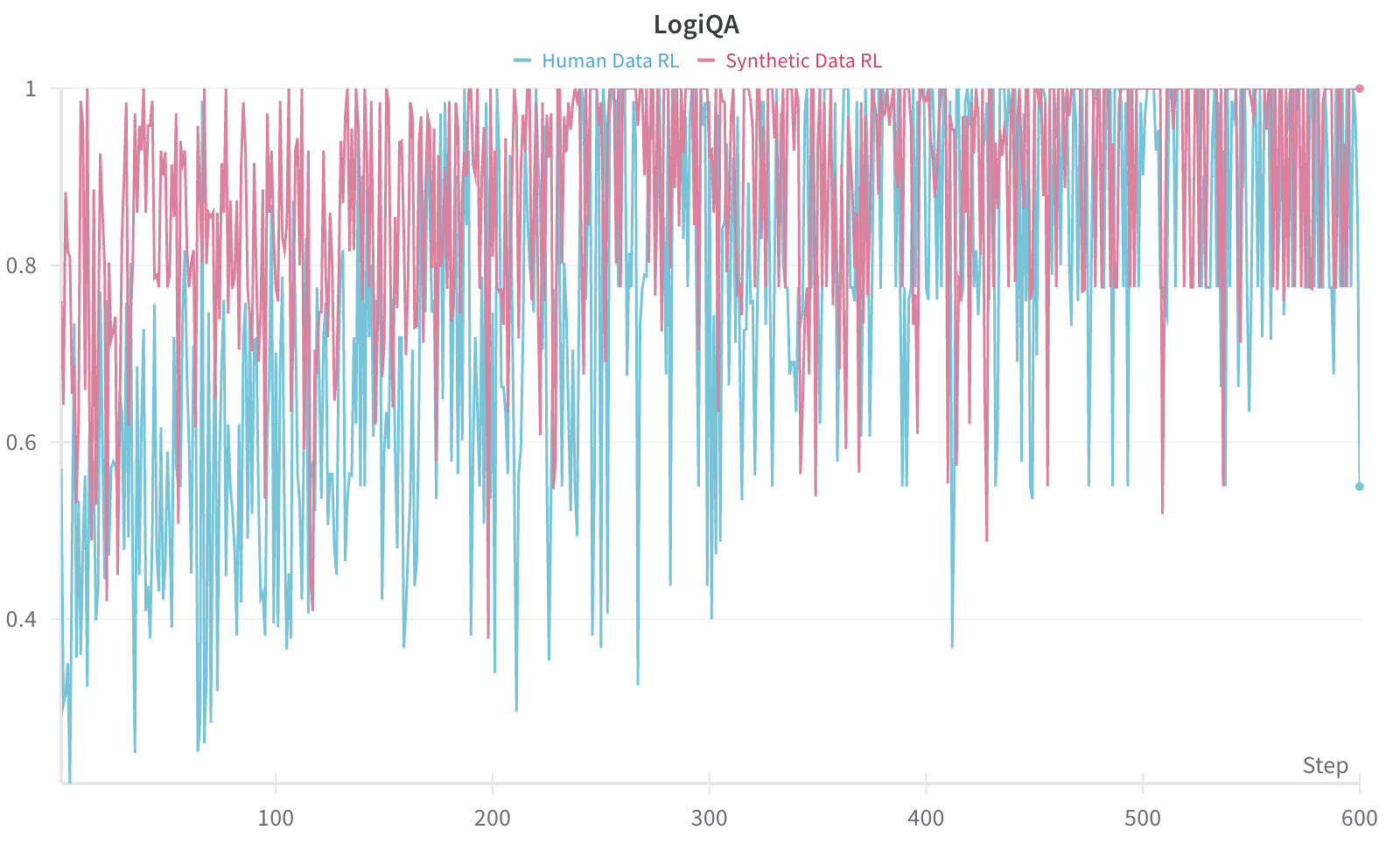}
        \caption{The training score (mean) curve on LogiQA}
        \label{fig:logiqa_score}
\end{figure}

\begin{figure} 
        \includegraphics[width=\linewidth]{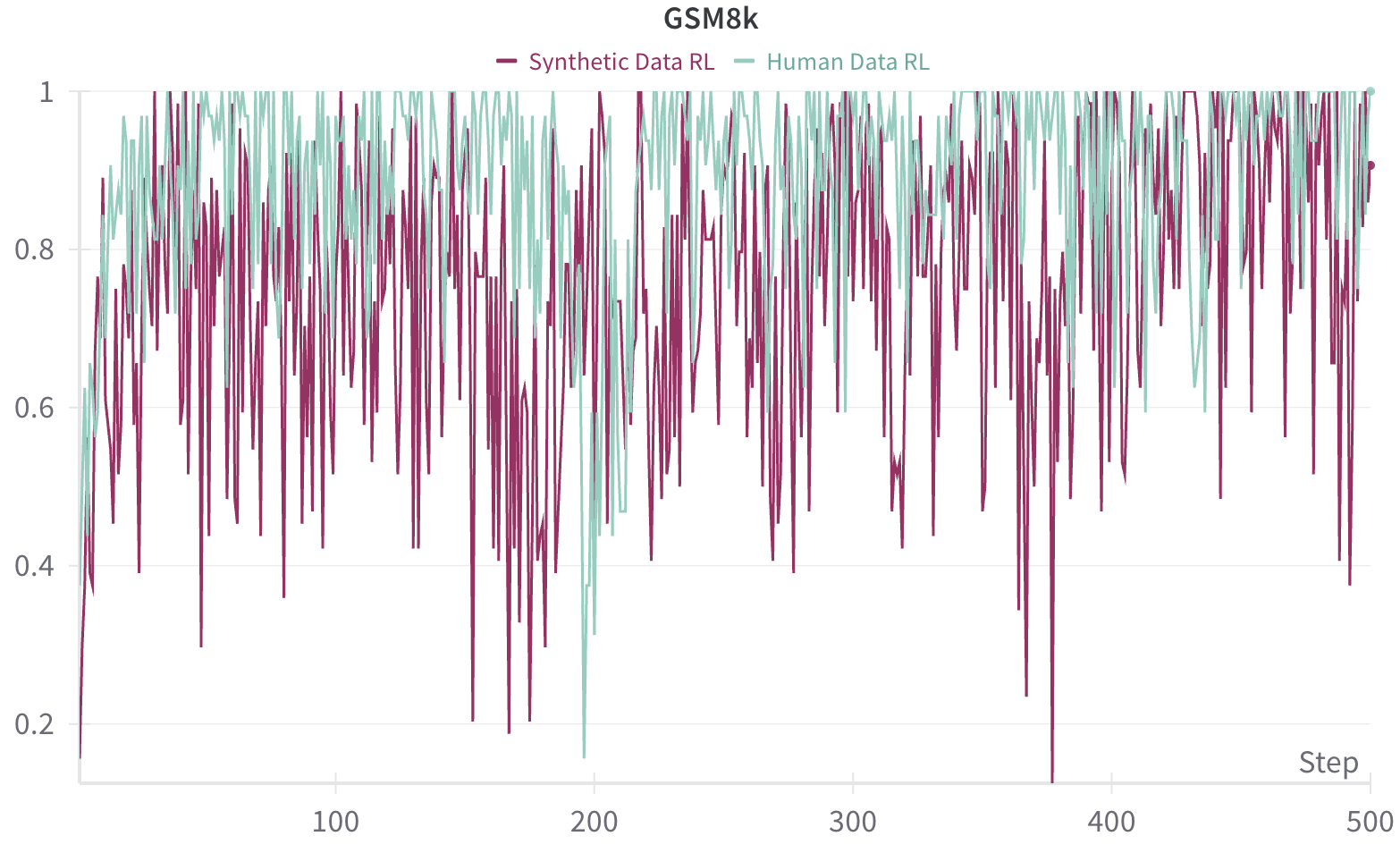}
        \caption{The training score (mean) curve on LogiQA}
        \label{fig:gsm8k_score}
\end{figure}

We analyze Synthetic Data RL training dynamics (e.g., KL loss, score, response length, test performance) and compare them to human data-based RL. While their learning curves differ notably, patterns vary across datasets—for instance, Synthetic Data RL outperforms on LogiQA (Figure~\ref{fig:logiqa_score}) but underperforms on GSM8K (Figure~\ref{fig:gsm8k_score}).
\subsection{Computational resources}
\label{appendix.compute}
We use 5 A100 GPUs to run our experiments. The average GPU hours for each experiment is 110.
\subsection{Limitations and societal impacts}
\label{appendix.limit}
This research has several limitations. Firstly, our study does not consider the complex multimodal settings or multi-turn agent tasks. Secondly, we did not focus on enhancing the GRPO reinforcement learning algorithm itself. Lastly, due to computational resource constraints, we were unable to evaluate the performance of larger models, such as a 14B parameter model, or explore the impact of a more extensive data budget, for example, 10,000 instances. These areas present opportunities for future investigation. As a machine learning work, we do not see any negative societal impact.

\end{document}